\documentclass[10pt,twocolumn,letterpaper]{article}

\usepackage[pagenumbers]{cvpr} % To force page numbers, e.g. for an arXiv version

\definecolor{cvprblue}{rgb}{0.21,0.49,0.74}
\usepackage[utf8]{inputenc}
\usepackage{newunicodechar}
\newunicodechar{（}{(}
\newunicodechar{）}{)}
\usepackage[pagebackref,breaklinks,colorlinks,allcolors=cvprblue]{hyperref}
\usepackage{xcolor}
\usepackage[table]{xcolor} 
\definecolor{mygreen}{HTML}{238b21}
\usepackage{hyperref}
\usepackage{xurl}
\usepackage{url}
\usepackage{caption}
\usepackage{float}
\usepackage{subcaption}
\usepackage{pifont}
\usepackage{makecell} 
\usepackage{multirow}
\usepackage{graphicx}
\usepackage{array}
\usepackage{wrapfig}
\usepackage{marvosym} % 添加marvosym包用于信封符号
\def\paperID{} % *** Enter the Paper ID here
\def\confName{CVPR}
\def\confYear{2026}

\title{Do All Individual Layers Help? An Empirical Study of Task\mbox{-}Interfering Layers in Vision-Language Models}

\author{Zhiming Liu$^{1}$ \;
Yujie Wei$^{2}$ \;
Lei Feng$^{3}$ \;
Xiu Su$^{4}$ \;
Xiaobo Xia$^{5}$ \;
Weili Guan$^{1}$ \; 
Zeke Xie$^{6}$ \;
Shuo Yang$^{1}$\textsuperscript{\Letter}\\
\\
[0.5em]
$^{1}$Harbin Institute of Technology, Shenzhen \quad
$^{2}$Harbin Institute of Technology \quad
$^{3}$Southeast University \\
$^{4}$Central South University \quad
$^{5}$National University of Singapore \\
$^{6}$The Hong Kong University of Science and Technology, Guangzhou\\[0.3em]
{\tt\small 2023312314@stu.hit.edu.cn, shuoyang@hit.edu.cn,}\\
{\small \url{https://mikuz12.github.io/Do_All_Individual_Layers_Help/}}\\
}
\vspace{-2cm}
\begin{document}
\maketitle
\begin{abstract}
Current Vision\mbox{-}Language Models (VLMs) have demonstrated remarkable capabilities across a wide range of multimodal tasks.
Typically, in a pretrained VLM, all layers are engaged by default to make predictions on downstream tasks. 
Surprisingly, we find that intervening on a single layer, such as by zeroing its parameters, can improve the performance on certain tasks, indicating that some layers hinder rather than help downstream tasks. To understand when and why this occurs, we systematically investigate how individual layers influence different tasks via layer intervention (e.g., parameter zeroing). Specifically, we measure the change in performance relative to the base model after intervening on each layer and observe improvements when bypassing specific layers. This improvement can be generalizable across models and datasets, indicating the presence of \textbf{Task\mbox{-}Interfering Layers} that harm downstream tasks' performance. To further analyze this phenomenon, we introduce \textbf{Task\mbox{-}Layer Interaction Vector}, which quantifies the effect of intervening on each layer of a VLM given a task. Crucially, these task\mbox{-}interfering layers exhibit task\mbox{-}specific sensitivity patterns: tasks requiring similar capabilities show consistent response trends under layer interventions, as evidenced by the high similarity in their task\mbox{-}layer interaction vectors. Inspired by these findings, we propose TaLo (\textbf{Ta}sk\mbox{-}Adaptive \textbf{L}ayer Kn\textbf{o}ckout) a training\mbox{-}free, test\mbox{-}time adaptation method that dynamically identifies and bypasses the most interfering layer for a given task as a means to validate and operationalize our observations. Without parameter updates, TaLo consistently improves performance across various models and datasets—even boosting Qwen–VL’s accuracy on the Maps task in ScienceQA by up to 16.6\%, serving as a proof-of-concept that demonstrates the tangible impact of this phenomenon. Our work reveals an unexpected form of modularity in pretrained VLMs and provides a plug\mbox{-}and\mbox{-}play, training\mbox{-}free mechanism to unlock hidden capabilities at inference time. The source code will be publicly available.

\end{abstract}
\section{Introduction}
\label{sec:intro}

Vision\mbox{-}Language Models (VLMs) have demonstrated remarkable success across diverse domains, including medicine~\citep{li2023llavamedtraininglargelanguageandvision,lin2025healthgptmedicallargevisionlanguage,yang2025medicallargevisionlanguage}, autonomous driving~\citep{sima2025drivelmdrivinggraphvisual, guo2024vlmautovlmbasedautonomousdriving}, and creative industries~\citep{wang2023docllmlayoutawaregenerativelanguage,huang2023languageneedaligningperception}, owing to their powerful cross\mbox{-}modal understanding and generation capabilities. 
In practical deployment, it is conventionally assumed that every layer in a VLM is actively utilized, thus justifying the use of the full model and requiring a complete computational pass during inference~\citep{Yin_2024}.
% A conventional assumption in their practical deployment is that all layers contribute positively or at least non-harmful to a given downstream task, thus necessitating a full computational pass during inference. 
However, our empirical investigation reveals a counterintuitive phenomenon: selectively bypassing a single layer of a pretrained model, can lead to substantial performance improvements on certain tasks.
This observation naturally leads to a fundamental question: \textit{\textbf{Do all individual layers in a pretrained VLM play a beneficial role in a specific task?}}
\begin{figure*}
    \centering
    \includegraphics[width=0.9\linewidth]{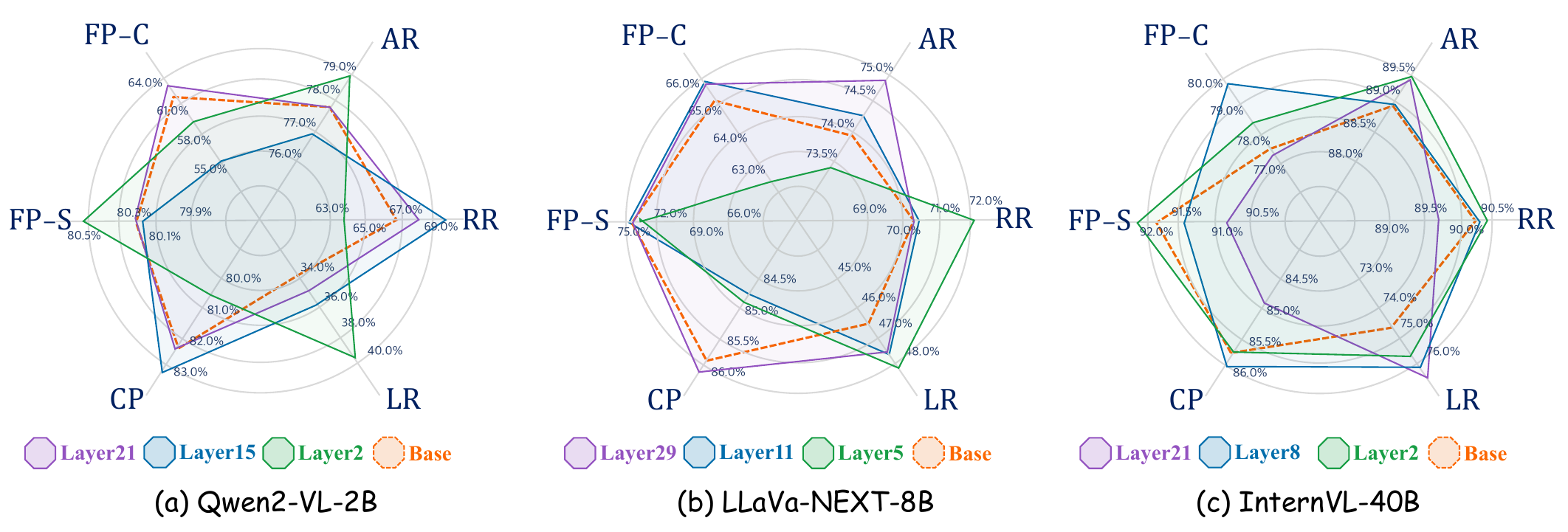}
    \caption{\textbf{Overview of the task\mbox{-}interfering layer phenomenon.} Each axis corresponds to a task category: AR (Attribute Reasoning), RR (Relation Reasoning), LR (Logical Reasoning), CP (Coarse Perception), FP\mbox{-}S (Fine\mbox{-}grained Perception [single\mbox{-}instance]), and FP\mbox{-}C (Fine\mbox{-}grained Perception [cross\mbox{-}instance]). Each plot shows model performance after zeroing out a single layer (solid curves), with the orange dashed line indicating the baseline performance (no intervention). In several tasks, performance improves upon layer removal, providing direct evidence for the existence of Task\mbox{-}Interfering Layers.}
    \vspace{-3mm}
    \label{overall}
\end{figure*}

To address the question, we first introduce layer intervention to quantify layer contributions towards a specific task: if performance on a task improves after intervening on a layer, we infer that the layer was previously hindering that task. Specially, we zero out the self\mbox{-}attention module of each layer, preserving residual connections while bypassing the attention mechanism, thereby nullifying the layer’s learned knowledge. As shown in Figure~\ref{overall}, zeroing specific layers leads to substantial performance gains on particular tasks across different models, while Figure~\ref{fig1-left} provides a more systematic and comprehensive analysis, illustrating that every task follows a characteristic pattern of performance change as different layers are intervened, reflecting task\mbox{-}specific sensitivities to layer functionality. We then introduce the \textbf{Task\mbox{-}Layer Interaction Vector} to further formalize this analysis. This vector quantifies the interaction between a task and each model layer by measuring performance changes under layer intervention. By encoding each task as such a vector, its unique sensitivity to layer manipulations becomes a computable and comparable representation. Building on this, a cluster analysis (see Figure~\ref{fig1-right}) of the correlations between these interaction vectors reveals that tasks requiring similar capabilities, such as mathematical reasoning, exhibit high similarity between their task\mbox{-}layer interaction vectors, indicating a strong relationship between the model's internal functional organization and the cognitive attributes of the tasks it performs, shedding light on the mechanisms behind its functional specialization.

While this phenomenon highlights the intricate functional organization within VLMs, a systematic exploration remains absent. Recent studies have noted that intervening layers (such as parameter zeroing or uniform scaling) in models can alter their general capabilities~\citep{zhang2024investigating,chen2025bringreasonvisionunderstanding}. However, this body of work has primarily highlighted the degradation of overall model performance, but overlooking the concurrent emergence of enhanced capabilities in certain downstream tasks. Crucially, existing research not only lacks a thorough understanding of this phenomenon but also overlooks its potential utility. Our work aims to address this critical gap. Our focus is not only on identifying what we term \textbf{Task\mbox{-}Interfering Layers}, which are layers whose presence actively constrains a model's potential on specific tasks, but also on uncovering the underlying mechanisms behind this phenomenon and exploring their practical applications.

Building on these motivations, we further introduce \textbf{TaLo} (\textbf{Ta}sk\mbox{-}Adaptive \textbf{L}ayer Kn\textbf{o}ckout), a minimalist, training\mbox{-}free framework designed to serve as a proof-of-concept for our observations. TaLo dynamically selects which layers to eliminate during inference for a given task, effectively enhancing its specific capabilities. The efficacy of this approach is validated across multiple VLMs and benchmarks. For LLaVA~\citep{li2024llavanext-strong}, applying the TaLo method yields up to a 4.7\% performance gain on the Tech\&Engineering task of MMMU~\citep{yue2023mmmu}. Remarkably, on the Physical Geography task of ScienceQA~\citep{lu2022learn}, it achieves an impressive 10.4\% improvement entirely without any parameter updates or additional training, serve as a clear validation that the Task-Interfering Layer phenomenon is not just statistically observable but also has practically significant implications.

We summarize our contributions as follows:

% contribution1:  Although layer-level modularity and redundancy have been previously observed, the possibility that some layers may actively impair performance on specific tasks remains underexplored. 
\begin{enumerate}
    % \item Through systematic parameter interventions, we uncover a counter\mbox{-}intuitive phenomenon where selectively skipping individual layers unlocks task\mbox{-}specific capabilities. To formalize this, we introduce the concept of \textbf{Task\mbox{-}Interfering Layers}, specialized components whose pretrained knowledge conflicts with the demands of certain tasks. 
    \item Through systematic layer-wise interventions, we observe that bypassing certain layers can lead to improved task performance. We refer to these as \textbf{Task-Interfering Layers}, denoting pretrained components whose pretrained knowledge are inconsistent with the objectives of specific downstream tasks.
    \item We establish a quantitative framework for analyzing the relationship between tasks and model layers by introducing the \textbf{Task\mbox{-}Layer Interaction Vector}, enabling further examination of how similar tasks exhibit consistent responses to layer interventions.
    \item We develop a practical, plug\mbox{-}and\mbox{-}play algorithm \textbf{TaLo} that leverages these insights to improve model performance at test time without any parameter updating. Using this method, LLaVA and Qwen\mbox{-}VL~\citep{wang2024qwen2} achieve peak improvements of up to 10.4\% and 16.6\%, respectively, from 10 tasks spanning 5 benchmarks.
\end{enumerate}

\section{Related Work}

Our research is situated at the intersection of Model Editing, Pruning, and Test\mbox{-}Time Adaptation (TTA). We draw upon concepts from model editing and pruning by using parameter intervention to modulate model behavior, yet we introduce a distinct approach focused on dynamic suppression for task\mbox{-}specific gains. Our method, TaLo, further contributes a novel paradigm to TTA by skipping the model's layer at inference time.

\subsection{Model Editing and Pruning}
\noindent{\textbf{Model Editing}} aims to modify pretrained models' behaviors or update knowledge without full retraining. Methods mainly fall into two categories. The first involves direct parameter updates, such as constrained fine\mbox{-}tuning to mitigate forgetting~\citep{zhu2020modifyingmemoriestransformermodels} or hyper\mbox{-}networks for dynamic parameter adjustment~\citep{decao2021editingfactualknowledgelanguage}. However, these are challenging to scale to large language models due to their parameter size. MEND~\citep{mitchell2022fastmodeleditingscale} addresses this by using low\mbox{-}rank gradient decomposition for efficient updates. The second category, Locate\mbox{-}and\mbox{-}Edit, identifies key parameters (e.g.,``knowledge neurons'') and applies targeted modifications~\citep{meng2023locatingeditingfactualassociations, dai2022knowledgeneuronspretrainedtransformers, meng2023masseditingmemorytransformer}. While enhancing interpretability, this approach is often labor\mbox{-}intensive and limited in scalability. Some methods instead maintain original parameters and use auxiliary modules for editing~\citep{mitchell2022memorybasedmodeleditingscale}. In multimodal settings, editing Vision\mbox{-}Language Models (VLMs) requires unique strategies. Directly porting LLM methods is ineffective; instead, recent work~\citep{chen2025attributionanalysismeetsmodel} proposes manipulating intermediate visual representations by identifying and editing regions most relevant to the target prompt, minimizing interference with unrelated features while preserving efficiency.

\noindent{\textbf{Model Pruning}} is distinct from model editing. It is a technique for model compression and acceleration that removes redundant or unimportant components, such as weights, neurons, or entire layers to reduce model size and improve inference speed while preserving performance~\citep{cheng2024surveydeepneuralnetwork,ma2020imageenhancingpatternbasedsparsity,he2017channelpruningacceleratingdeep,dumitru2024layer,muralidharan2024compact,siddiqui2024deeper,yin2023outlier}. Its primary goal is efficiency optimization, not enhancing or correcting the model’s knowledge.

In contrast, our method, TaLo, aims to improve task performance by temporarily suppressing harmful reasoning pathways during inference, without permanently altering the model. This differs fundamentally from both model editing and pruning. Notably, TaLo circumvents gradient-based fine-tuning or fine-grained parameter manipulation\citep{zheng2023regularized,ali2025detecting}, which are computationally expensive and inefficient at inference time, by operating instead at the block level through reversible, test-time interventions. Our results demonstrate that performance gains can be achieved not only by adding or removing model components, but also by strategically inhibiting existing ones during inference.

\subsection{Test\mbox{-}Time Adaptation}
\label{tta}
Test\mbox{-}Time Adaptation~(TTA) aims to dynamically adjust models to shifting data distributions during inference, a crucial step for robust deployment in real\mbox{-}world scenarios. Prevailing approaches either update model components like weights or normalization statistics using test batches~\citep{iwasawa2021testtime,wang2021tentfullytesttimeadaptation,yi2023temporalcoherenttesttimeoptimization,sun2020testtimetrainingselfsupervisiongeneralization,schneider2020improvingrobustnesscommoncorruptions}, or, particularly for vision\mbox{-}language models, fine\mbox{-}tune learnable prompts~\citep{zhang2022memotesttimerobustness,shu2022testtimeprompttuningzeroshot,feng2023diversedataaugmentationdiffusions}. Other methods~\citep{karmanov2024efficienttesttimeadaptationvisionlanguage} perform zero\mbox{-}shot classification using test\mbox{-}time feature caching. Our work introduces a distinct TTA paradigm based on layer intervention. Guided by a small set of unlabeled test samples, TaLo identifies a single, fixed layer who is harmful for the target task and zeroes it out for all subsequent instances of that task. Crucially, this adaptation is task-level and static: once the intervention is determined, no per-instance decisions or parameter updates are needed during inference.
\begin{figure*}[t!]
    \centering
    \begin{subfigure}[h]{0.48\linewidth}
        \centering
        \includegraphics[width=\linewidth]{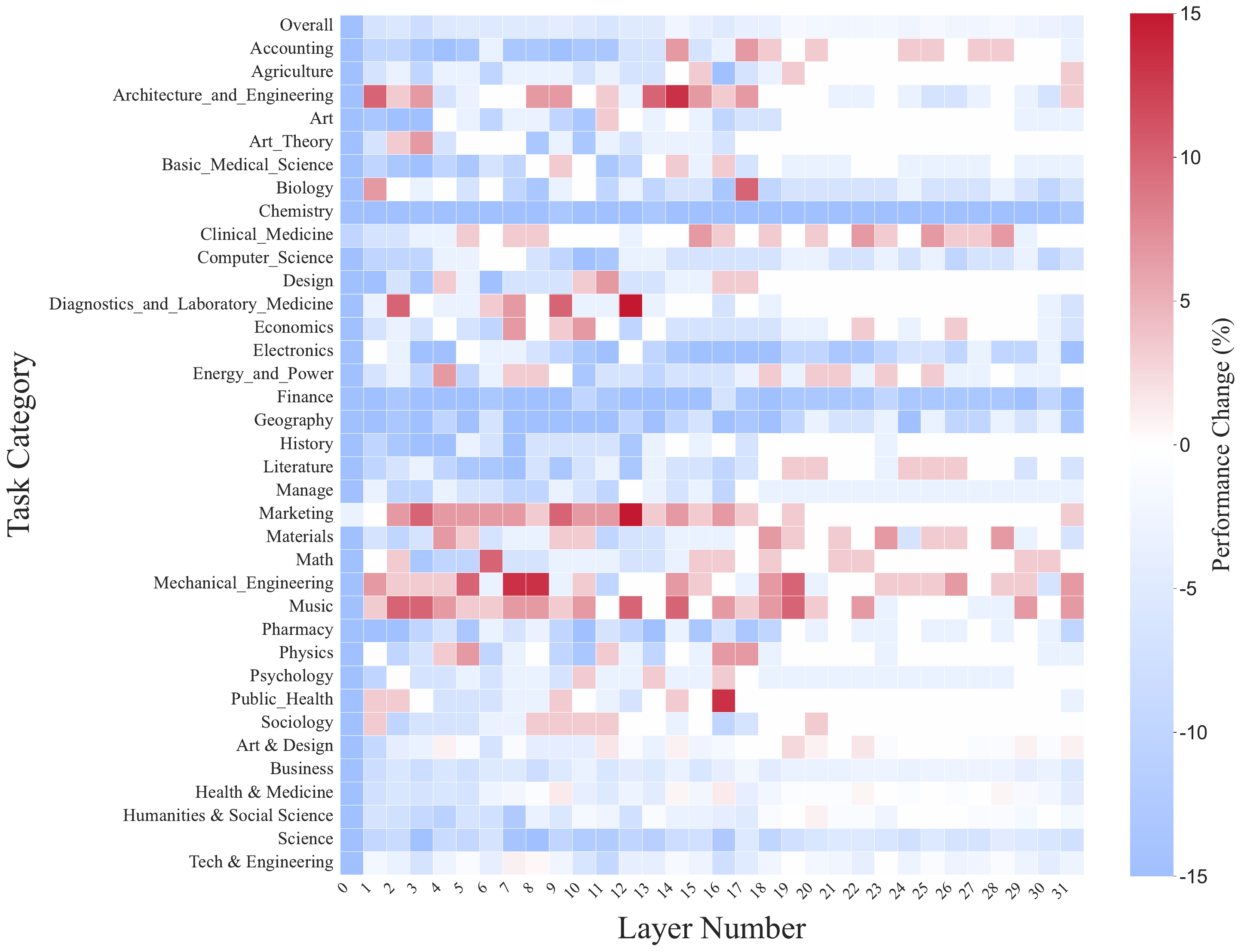}
        \caption{}
        \label{fig1-left}
    \end{subfigure}
    \hfill
    \begin{subfigure}[h]{0.42\linewidth}
        \centering
        \includegraphics[width=\linewidth]{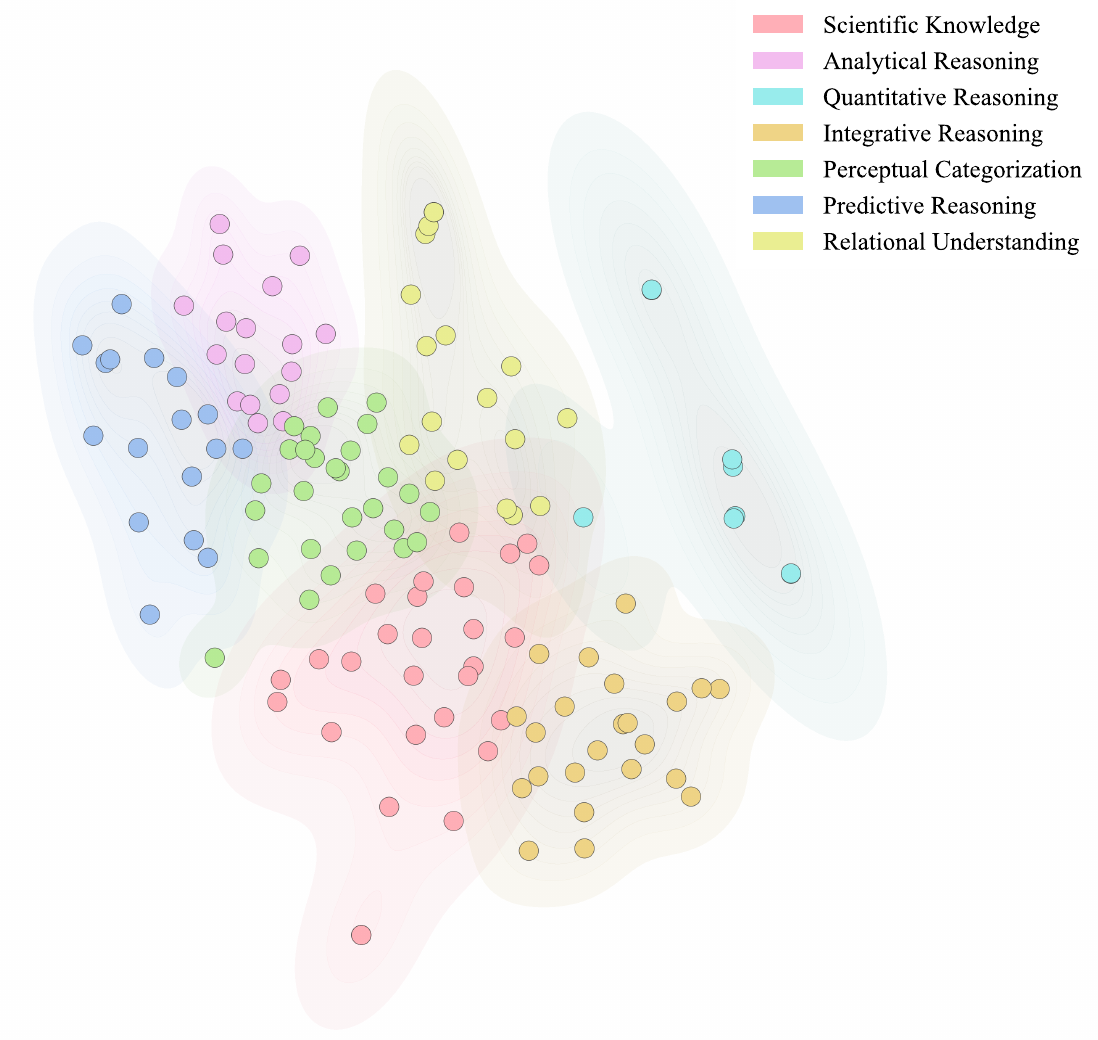}
        \caption{}
        \label{fig1-right}
    \end{subfigure}
    \caption{\textbf{Empirical Validation of the Task\mbox{-}Interfering Layers.} 
        \textbf{(a)} Visualization of the percentage change in accuracy across tasks after zeroing each layer on \textit{LLaVA\mbox{-}Next\mbox{-}LLaMA3\mbox{-}8B}. Red indicates performance improvements relative to the base model, while blue indicates degradation. Many tasks show performance gains under layer interventions, indicating that interfering layers are commonly exist in VLMs.
        \textbf{(b)} The t\mbox{-}SNE visualization of task clusters. Each point in the figure represents a task encoded as a Task\mbox{-}Layer Interaction Vector $\mathbf{v}^{(\mathcal{T})}$. Tasks are clustered based on their pairwise similarities measured by Pearson correlation $\rho_{ij}$ between different vectors, with tasks requiring similar capabilities forming coherent clusters. The color\mbox{-}coded clusters correspond to different types of tasks, indicating that tasks with shared cognitive demands exhibit similar intervention responses, reflecting a structured functional layout in LLaVA~\citep{li2024llavanext-strong} 
        (See Table~\ref{tab:clusters_tasks} for complete clustering details).
    }
    \label{fig1}
\end{figure*}
This training\mbox{-}free strategy avoids the large\mbox{-}scale parameter updates and complex prompt modifications inherent in prior methods. Consequently, it enables efficient, plug\mbox{-}and\mbox{-}play adaptation with minimal computational overhead, as the original model parameters remain intact and reusable across tasks.

\section{Discovering and Characterizing Task-Interfering Layers}
 
We begin by using parameter intervention to uncover the commonly existing phenomenon of Task\mbox{-}Interfering Layers. Building on this, we introduce the Task\mbox{-}Layer Interaction Vector, which enables a more systematic analysis, effectively uncovering patterns and revealing the coherence of layer\mbox{-}level interference.

\subsection{Discovering Task\mbox{-}Interfering Layers}

Our aim is to isolate and quantify the contribution of individual layers to specific task capabilities. To achieve this, we employ parameter intervention to systematically probe the functional role of each layer inspired by~\citet{zhang2024investigating, chen2025bringreasonvisionunderstanding}. Specifically, for each layer, we replace the parameters of the self\mbox{-}attention module with zeros or a uniform distribution (i.e., setting every parameter to an identical value $1/N$, where $N$ corresponds to the matrix's first dimension) and evaluate the resulting change in task accuracy against the unmodified model. For parameter zeroing, it effectively nullifies the attention mechanism, leaving only the residual connection, which facilitates direct communication between distant layers, bypassing intermediate transformations. As for Uniform Scaling, it reduces the complex attention operation to a simple global averaging of the input features, causing the output to become a rank\mbox{-}one matrix. Our hypothesis is that a performance increase after intervention suggests that the layer was hindering task performance,  indicating its role as a Task\mbox{-}Interfering Layer. Conversely, a performance drop indicates that the layer contributed positively to the task.

We then apply parameter zeroing intervention to LLaVA\mbox{-}Next~\citep{li2024llavanext-strong}, a model consisting of 32 layers, and evaluate performance on the MMMU~\citep{yue2023mmmu} dataset. As shown in Figure~\ref{fig1-left}, 54.1\% of tasks exhibit performance gains exceeding 5\% when a single layer's parameters are zeroed. Similar trends are observed on other models and datasets: for Qwen\mbox{-}VL, the proportion reaches 75.6\%. To further validate the generality of our findings, we present additional experiments on diverse models, benchmarks, and intervention strategies in Appendix~\ref{results}. This consistent pattern provides direct empirical evidence for the existence of Task\mbox{-}Interfering Layers whose activation hinders rather than helps task performance.

It is worth noting that our study focuses specifically on intervening in the LLM backbone of vision–language models. The visual encoder (e.g., a CLIP-based ViT) is treated as a fixed feature extractor and left unmodified, as it is already highly optimized for visual representation. We hypothesize that the primary source of task interference lies not in perception but in cross-modal reasoning and task execution—processes predominantly handled by the LLM component. Also, the design choice to intervene only in self-attention modules is deliberate: preliminary experiments show that applying the same zeroing intervention to feed-forward network modules, even in a single layer, leads to catastrophic performance degradation and often produces semantically incoherent outputs. Consequently, throughout this work, ``layer intervention'' refers exclusively to modifying the self-attention module within a given layer.
\begin{figure*}
    \centering
    \includegraphics[width=0.8\linewidth]{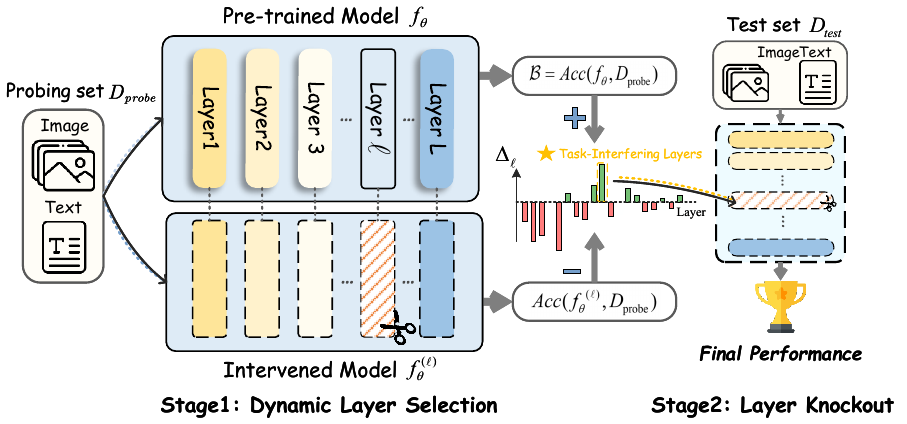}
    \caption{\textbf{Framework of TaLo.} TaLo first dynamically selects the Task\mbox{-}Interfering layer for a specific task and knocks out that layer in the final evaluation procedure.}
    \label{fig:talo}
\end{figure*}

\vspace{-0.5em}

% \subsection{Characterizing Task-Interfering Layers through modeling task-layer interaction}
\subsection{Characterizing Task\mbox{-}Interfering Layers}
\paragraph{Modeling Task\mbox{-}Layer Interaction into Vector Space.}
To uncover the systematic response patterns of tasks to layer interventions, we introduce the Task\mbox{-}Layer Interaction Vector. This vector is designed to model the relationship between a task's performance and interventions applied to each layer of the model.

Specifically, the Task\mbox{-}Layer Interaction Vector is a representation that characterizes a task's sensitivity to each model layer. Each dimension of the vector corresponds to a network layer and captures the change in task accuracy caused by intervening on that layer relative to the base model. A positive value indicates that the layer interferes with the task, which manifests as an improvement in accuracy upon intervention. Conversely, a negative value indicates that the layer contributes positively to the task, reflected in a drop in accuracy when the layer is modified. More formally, for task $\mathcal{T}$, and a model with $L$ layers, the \textbf{Task\mbox{-}Layer Interaction Vector} is defined as:
\begin{equation}
    \mathbf{v}^{(\mathcal{T})} = \left( v_1^{(\mathcal{T})}, v_2^{(\mathcal{T})}, \dots, v_L^{(\mathcal{T})} \right) \in \mathbb{R}^L,
\end{equation}
where each element $v_i^{(\mathcal{T})}$, referred to as the \textbf{layer sensitivity score}, quantifies the change in task performance upon intervention at layer $i$. Formally, it is defined as
\begin{equation}
    v_i^{(\mathcal{T})} = \text{Acc}\left( \mathcal{M}_{\text{intv}}^{(i)},\mathcal{T} \right) - \text{Acc}\left( \mathcal{M}_{\text{base}},\mathcal{T} \right).
\end{equation}
Here, $\text{Acc}(\cdot,\mathcal{T})$ denotes the accuracy on task $\mathcal{T}$, $\mathcal{M}_{\text{base}}$ is the base model without intervention, and $\mathcal{M}_{\text{intv}}^{(i)}$ is the model with the $i$\mbox{-}th layer's parameters intervened.

Through this vector representation, we abstract the influence of each layer on a task into a structured representation within a high\mbox{-}dimensional vector space. This provides a quantifiable and comparable analytical tool for subsequent pattern analysis.

\paragraph{Characterizing Task\mbox{-}Layer Interaction Patterns.}

We assume that tasks drawing upon the same underlying cognitive skills (e.g., numerical reasoning or arithmetic reasoning) should engage similar internal processing pathways within the model. Since Task\mbox{-}Layer Interaction Vector $\mathbf{v}^{(\mathcal{T})}$ captures a task’s dependence on each layer, reflecting how it propagates through the model’s architecture. We hypothesize that related tasks will exhibit highly correlated interaction vectors.

To validate this hypothesis, we conduct a systematic analysis across 6 benchmarks and nearly 100 tasks. For each pair of tasks, $\mathcal{T}_i$ and $\mathcal{T}_j$ (hereafter, we use indices $i$ and $j$ as shorthand for the corresponding tasks in this section), we compute their Pearson correlation coefficient, $\rho_{ij} = \text{Corr}(\mathbf{v}^{(i)}, \mathbf{v}^{(j)})$, and define a distance metric as $d_{ij} = 1 - \rho_{ij}$. This ensures that more similar tasks have a smaller distance, providing a solid basis for clustering.

As shown in Figure~\ref{fig1-right}, the results confirm our hypothesis: tasks that rely on shared abilities cluster together in the task\mbox{-}layer interaction space, reflecting their common internal processing mechanisms. This suggests that tasks sharing underlying cognitive or domain\mbox{-}specific demands exhibit highly similar sensitivity patterns to layer interventions, revealing a structured organization of functional dependencies across the model’s architecture. For instance, one prominent cluster is dominated by \textbf{quantitative reasoning} tasks (e.g., \textit{numeric commonsense}, \textit{arithmetic reasoning}, and \textit{geometry}). Another distinct cluster groups together domain\mbox{-}specific \textbf{scientific tasks} such as \textit{Physics} and \textit{Scientific Reasoning}, reflecting their shared reliance on formal scientific knowledge. This demonstrates a strong alignment between the model’s internal response to layer interventions 
and the underlying cognitive structure of tasks, revealing that task\mbox{-}interfering layers generalize across tasks with similar abilities. This generalization suggests that layer sensitivity is determined by functional demands, not task\mbox{-}specific properties, enabling reliable estimation of interfering layers from a few representative samples.

\paragraph{Consistency across Intervention Methods.}
\label{consistency}
\begin{figure}
    \centering
    \includegraphics[width=0.75\linewidth]{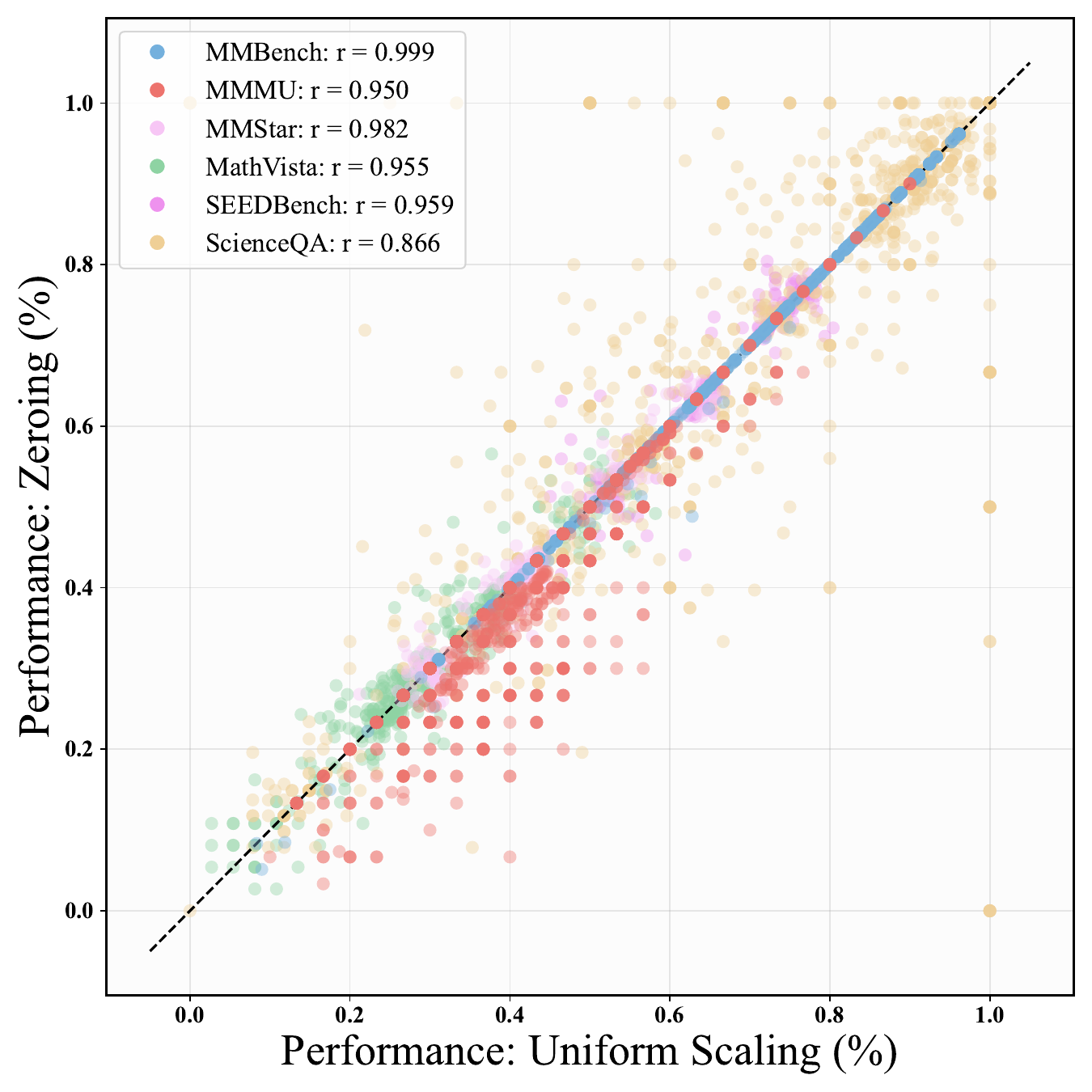}
    \caption{Consistency analysis of different interventions.}
    \label{fig:vs}
\end{figure}
To validate that the Task\mbox{-}Interfering Layer phenomenon is not a methodological artifact, we examine the consistency of findings across two distinct intervention strategies. For each task $\mathcal{T}$ and layer $i$, we measure the model's accuracy under both intervention types, yielding two performance scores: $\text{Acc}(\mathcal{M}_{\text{zero}}^{(i)}, \mathcal{T})$ and $\text{Acc}(\mathcal{M}_{\text{unif}}^{(i)}, \mathcal{T})$. We perform this analysis across six benchmarks. To visualize the relationship, we generate scatter plots (Figure~\ref{fig:vs}) where each point represents a single task\mbox{-}layer pair, with its coordinates determined by the accuracies under uniform scaling (x\mbox{-}axis) and parameter zeroing (y\mbox{-}axis). To quantify the level of agreement, we then compute the Pearson correlation coefficient across all points for each benchmark. The results reveal a strong and statistically significant positive correlation across all benchmarks. This reproducibility across fundamentally different intervention strategies strongly reinforces the validity of our findings. It affirms that the existence of Task\mbox{-}Interfering Layers is not an artifact of a specific intervention choice, but rather reflects an intrinsic property of the model arising from task conflicts during pretraining.
\section{Task\mbox{-}Adaptive Layer Knockout}
Building upon the discovery of Task\mbox{-}Interfering Layers in VLMs, and inspired by advances in dynamic model adaptation~\citep{wang2018skipnet,zhao2025skipgptdynamiclayerpruning,cao2024mentored,sun2025transformersquaredselfadaptivellms,cao2023data}, we propose a simple yet effective algorithm—\textbf{Ta}sk\mbox{-}Adaptive \textbf{L}ayer Kn\textbf{o}ckout \textbf{(TaLo)}, a training\mbox{-}free method for task\mbox{-}level model customization at test time which serves as a proof of concept to validate both the existence and practical utility of Task-Interfering Layers. TaLo operates in two stages: (1) dynamic layer selection for a specific task, and (2) task\mbox{-}interfering layer knockout on the model. This approach enables performance enhancement on specific tasks by strategically exploiting the task\mbox{-}interfering layers within the model.

Our approach is guided by the principle of targeted intervention. The objective is to identify an optimal layer whose modification unlocks a model’s latent, task\mbox{-}specific abilities. The search begins by establishing a performance baseline. For an $L$-layer model $f_\theta$ with parameters $\theta = \{\theta_1, ..., \theta_L\}$ and a probing set $D_{\text{probe}} = \{(x_i, y_i)\}_{i=1}^N$ sampled from a given downstream task in an $N$-shot setting, we define the baseline score $\mathcal{B}$ on the unmodified model: $ \mathcal{B} = {Acc}(f_\theta, D_{\text{probe}})$.
In our experiments, ${Acc}$ is accuracy on probing set $D_{\text{probe}}$. Then, we systematically test each layer's potential. For each layer $\ell$, we apply an intervention $I$ (Here, we utilize parameter zeroing, $I(\theta_\ell)=0$) to create a modified model $f_{\theta}^{(\ell)}$ and measure the resulting accuracy gain $\Delta_\ell$:
\begin{equation}
    \Delta_\ell = {Acc}(f_{\theta}^{(\ell)}, D_{\text{probe}}) - {Acc}(f_\theta, D_{\text{probe}}).
\end{equation}
This iterative process reveals how each layer's function contributes to the specific task. The search concludes when we identify the optimal layer $\ell^*$ responsible for the maximal positive performance gain:
\begin{equation}
    \ell^* = \mathop{\text{argmax}}\limits_{\ell \in \{1,...,L\}} \Delta_\ell.
\end{equation}
This layer is then selected as the task\mbox{-}interfering layer. If no such layer exhibits a statistically significant sensitivity peak, we retain the original model without any modification. Having identified the layer, we proceed to apply the intervention (i.e., knockout) to this layer during inference. We then evaluate the intervened model on the held\mbox{-}out test samples of the target task, which are those not included in the initial probe set, to ensure an unbiased assessment.

\begin{table*}[t!]
\centering
\renewcommand{\arraystretch}{1.3} % 调整行距
\small
\caption{The performance of \textbf{TaLo} applied to two models (Zeroing): \textit{LLaVA\mbox{-}Next\mbox{-}LLaMA3\mbox{-}8B}, \textit{Qwen2\mbox{-}VL\mbox{-}2B} across five benchmarks. MMStar (\textit{Logical reasoning} annotated as L\mbox{-}R), MMBench (\textit{Physical property reasoning} annotated as PP\mbox{-}R, \textit{image emotion} annotated as I-E), MMMU (\textit{Tech\&Engineering} annotated as T\&E, \textit{Health\&Medicine} annotated as H\&M), ScienceQA (\textit{Physical Geography} annotated as P\mbox{-}G), and SEEDBench (\textit{Visual Reasoning} annotated as V\mbox{-}R, \textit{Text Understanding} annotated as T\mbox{-}U). Additional results for TaLo on various tasks are provided in Table~\ref{tab:llava} and Table~\ref{tab:qwen}. Shot number indicates probe set size. Base performance may vary across tasks, especially those with limited samples under different shot settings. To ensure a fair evaluation, we therefore focus primarily on the relative improvement over the base model.}
\scalebox{0.8}{

\begin{tabular}{l|l|ll|ll|ll|ll|ll|c}
\toprule[1.2pt]
\multicolumn{1}{l|}{\multirow{2}{*}{\textbf{Model}}} & \multirow{2}{*}{\textbf{Shots}} 
& \multicolumn{2}{c|}{\textbf{MMStar}} 
& \multicolumn{2}{c|}{\textbf{MMBench}} 
& \multicolumn{2}{c|}{\textbf{MMMU}} 
& \multicolumn{2}{c|}{\textbf{ScienceQA}} 
& \multicolumn{2}{c|}{\textbf{SEEDBench}} 
& \multirow{2}{*}{\textbf{Avg}} \\ 
\cmidrule{3-4} \cmidrule{5-6} \cmidrule{7-8} \cmidrule{9-10} \cmidrule{11-12}
& &\textbf{Math} &\textbf{I-R} &\textbf{PP-R} &\textbf{I-E} &\textbf{T\&E} &\textbf{H\&M} &\textbf{P-G} &\textbf{Maps} &\textbf{V-R} &\textbf{T-U} & \\ 
\midrule
\multicolumn{1}{l|}{\multirow{3}{*}{\textbf{LLaVA}}} 
& \textbf{10 shots} &32.9 \textcolor{mygreen}{\scriptsize$\uparrow$2.5} &58.3 \textcolor{mygreen}{\scriptsize$\uparrow$4.2}&55.3 \textcolor{mygreen}{\scriptsize$\uparrow$7.8} &65.6 \textcolor{mygreen}{\scriptsize$\uparrow$0.6} &35.2 \textcolor{mygreen}{\scriptsize$\uparrow$1.1} &42.2 \textcolor{mygreen}{\scriptsize$\uparrow$5.1} &34.5 \textcolor{mygreen}{\scriptsize$\uparrow$6.9} &16.7 \textcolor{mygreen}{\scriptsize$\uparrow$2.4} &70.4 \textcolor{mygreen}{\scriptsize$\uparrow$1.3} &53.6 \textcolor{mygreen}{\scriptsize$\uparrow$7.2} &\textcolor{mygreen}{3.91$\uparrow$} \\
\multicolumn{1}{l|}{} & \cellcolor{gray!15}\textbf{15 shots}  &\cellcolor{gray!15}33.2 \cellcolor{gray!15}\textcolor{mygreen}{\scriptsize$\uparrow$3.0} &\cellcolor{gray!15}57.9 \textcolor{mygreen}{\scriptsize$\uparrow$4.3} &\cellcolor{gray!15}56.0 \textcolor{mygreen}{\scriptsize$\uparrow$3.8} &\cellcolor{gray!15}69.3 \textcolor{mygreen}{\scriptsize$\uparrow$0.7} &\cellcolor{gray!15}37.5 \textcolor{mygreen}{\scriptsize$\uparrow$8.3} &\cellcolor{gray!15}35.3 \textcolor{mygreen}{\scriptsize$\uparrow$5.9} &\cellcolor{gray!15}\textbf{41.4 \textcolor{mygreen}{\scriptsize$\uparrow$10.4}} &\cellcolor{gray!15}23.8 \textcolor{mygreen}{\scriptsize$\uparrow$7.1} &\cellcolor{gray!15}70.8 \textcolor{mygreen}{\scriptsize$\uparrow$1.1} &\cellcolor{gray!15}55.4 \textcolor{mygreen}{\scriptsize$\uparrow$9.0} &\cellcolor{gray!15}\textcolor{mygreen}{5.36$\uparrow$} \\
\multicolumn{1}{l|}{} & \textbf{20 shots} &33.9 \textcolor{mygreen}{\scriptsize$\uparrow$3.0} &58.7 \textcolor{mygreen}{\scriptsize$\uparrow$4.3} &53.4\textcolor{mygreen}{\scriptsize$\uparrow$4.8} &66.4\textcolor{red}{\scriptsize$\downarrow$1.5} &29.5 \textcolor{red}{\scriptsize$\downarrow$3.6} &46.2\textcolor{gray}{\scriptsize$-$0.0} &34.5 \textcolor{mygreen}{\scriptsize$\uparrow$3.5} &26.2 \textcolor{mygreen}{\scriptsize$\uparrow$9.5} &72.5 \textcolor{mygreen}{\scriptsize$\uparrow$2.0} &55.4 \textcolor{mygreen}{\scriptsize$\uparrow$5.4} &\textcolor{mygreen}{2.74$\uparrow$} \\ 
\midrule

\multicolumn{1}{l|}{\multirow{3}{*}{\textbf{Qwen-VL}}} 
&\cellcolor{gray!15} \textbf{10 shots} 
&\cellcolor{gray!15}44.8 \textcolor{gray}{\scriptsize$-$0.0} 
&\cellcolor{gray!15}54.8 \textcolor{red}{\scriptsize$\downarrow$1.9} &\cellcolor{gray!15}48.6 \textcolor{mygreen}{\scriptsize$\uparrow$8.5} 
&\cellcolor{gray!15}69.4 \textcolor{mygreen}{\scriptsize$\uparrow$1.3} 
&\cellcolor{gray!15}28.6 \textcolor{mygreen}{\scriptsize$\uparrow$0.6} 
&\cellcolor{gray!15}34.5 \textcolor{red}{\scriptsize$\downarrow$1.6} &\cellcolor{gray!15}31.0\textcolor{gray}{\scriptsize$-$0.0} 
&\cellcolor{gray!15}31.0 \textcolor{mygreen}{\scriptsize$\uparrow$2.4} 
&\cellcolor{gray!15}70.8 \textcolor{red}{\scriptsize$\downarrow$1.5} 
&\cellcolor{gray!15}57.1 \textcolor{mygreen}{\scriptsize$\uparrow$8.9} &\cellcolor{gray!15}\textcolor{mygreen}{1.67$\uparrow$} \\

\multicolumn{1}{l|}{} & \textbf{15 shots} &44.2 \textcolor{mygreen}{\scriptsize$\uparrow$0.5} &54.2 \textcolor{red}{\scriptsize$\downarrow$1.6} &55.3 \textcolor{mygreen}{\scriptsize$\uparrow$1.2} &65.7 \textcolor{red}{\scriptsize$\downarrow$0.7} &24.4 \textcolor{red}{\scriptsize$\downarrow$1.9} &35.8 \textcolor{red}{\scriptsize$\downarrow$2.9} &34.5 \textcolor{mygreen}{\scriptsize$\uparrow$3.5} &\textbf{45.2 \textcolor{mygreen}{\scriptsize$\uparrow$16.6}} &71.2 \textcolor{red}{\scriptsize$\downarrow$2.6} &55.4 \textcolor{mygreen}{\scriptsize$\uparrow$1.8} &\textcolor{mygreen}{1.39$\uparrow$} \\

\multicolumn{1}{l|}{} & \textbf{20 shots} &38.2 \textcolor{red}{\scriptsize$\downarrow$1.2} &55.3 \textcolor{red}{\scriptsize$\downarrow$0.5}&60.3 \textcolor{mygreen}{\scriptsize$\uparrow$4.8} &67.9 \textcolor{mygreen}{\scriptsize$\uparrow$0.7} &26.1 \textcolor{red}{\scriptsize$\downarrow$0.5}& 23.5 \textcolor{red}{\scriptsize$\downarrow$5.9}&37.9\textcolor{gray}{\scriptsize$-$0.0} &16.7\textcolor{gray}{\scriptsize$-$0.0} &71.3\textcolor{red}{\scriptsize$\downarrow$0.8} & 44.6 \textcolor{gray}{\scriptsize$-$0.0}&\textcolor{red}{0.34$\downarrow$} \\ 
\bottomrule[1.1pt]
\end{tabular}
}

\label{tab1}
\end{table*}

\section{Experiments}
\subsection{Setups}
\noindent{\textbf{Models and Benchmarks.}} We conducted experiments on three VLMs of varying scales: Qwen2\mbox{-}VL\mbox{-}2B~\citep{wang2024qwen2}, LLaVA\mbox{-}Next\mbox{-}LLaMA3\mbox{-}8B, and InternVL2\mbox{-}26B~\citep{chen2024internvl}. To assess the impact of our interventions, we evaluated both the intervened and original models across five key multiple\mbox{-}choice question (MCQ) benchmarks: MMStar~\citep{chen2024we}, MMBench~\citep{liu2024mmbench}, MMMU~\citep{yue2023mmmu}, ScienceQA~\citep{lu2022learn}, and SEEDBench~\citep{li2023seed}. Further details on the benchmarks and model configurations are provided in Appendix~\ref{Models and Benchmarks}. 

\noindent{\textbf{Baselines.}} We aim to achieve task\mbox{-}specific gains via training\mbox{-}free, plug\mbox{-}and\mbox{-}play layer intervention. All comparisons are against the original pretrained model. In Section~\ref{analysis}, we also evaluate fine\mbox{-}tuning methods using the same few\mbox{-}shot samples as TaLo. It is critical to note that the baseline performance reported in our tables reflects the accuracy on these specific tasks. This performance may differ substantially from the overall average accuracy reported for an entire benchmark. We verified our baseline implementations using the official evaluation harness and standard chat templates to ensure a fair and accurate comparison.

\noindent{\textbf{Task Selection Rationale. }} Rather than reporting average performance across entire benchmark suites which risks obscuring task-specific behaviors, we selectively analyze individual sub-tasks. We find that not all tasks exhibit what we term ``task-interference layers.'' As evidenced by the sensitivity visualizations in Figure~\ref{fig1-left} and the Appendix~\ref{results}, the heatmaps for certain tasks are predominantly blue, indicating that intervening on any layer results in performance degradation. This confirms that for these specific cases, no significant interference exists to be mitigated, rendering our method naturally inapplicable. Therefore, we focus specifically on those tasks where the pretrained model performs poorly, as they often reveal biases or gaps in the model’s pretrained knowledge. The goal of this work is not to claim universal performance gains across all tasks, but rather to provide a proof of concept (as discussed in Section~\ref{disscussion}) that layer-level interventions can unlock latent capabilities in these particular weak spots of the model.

\noindent{\textbf{Implementation Details. }}  All experiments were conducted on a single \textit{80GB A100 GPU} under identical environments, ensuring reproducibility and fair comparisons. Evaluations used the standardized VLMEvalKit framework~\citep{duan2024vlmevalkit}. For more experimental details, please refer to the Appendix~\ref{appendix:experimental_details}.

\begin{table*}[htpb]
\centering
\renewcommand{\arraystretch}{1.2} % 调整行距
\small
\caption{The performance of \textbf{TaLo} applied to \textit{InternVL2-26B} (Zeroing) across multiple tasks. \textit{Persuasive strategies} annotated as P\mbox{-}S, \textit{Basic economic principles} annotated as B\mbox{-}EP, \textit{Physical Geography} annotated as P\mbox{-}G, \textit{Geography} annotated as Geo, The Americas: geography annotated as A:Geo, and \textit{G2T} annotated as Genes to traits.}
\scalebox{0.85}
{
\begin{tabular}{l|l|llllllll|l}
\toprule[1.2pt]
\multicolumn{1}{l|}{\multirow{2}{*}{\textbf{Model}}} & \multirow{2}{*}{\textbf{Shots}} 
& \multicolumn{8}{c|}{\textbf{Tasks}} & \multirow{2}{*}{\textbf{Avg}} \\ 
\cmidrule{3-10}
& &\textbf{P-S} &\textbf{B-EP} &\textbf{P-G} &\textbf{Solutions} &\textbf{Geo} &\textbf{A:Geo} &\textbf{G2T} &\textbf{Materials} & \\ 
\midrule
\multicolumn{1}{l|}{\multirow{2}{*}{\textbf{InternVL}}} 
&\cellcolor{gray!15} \textbf{10 shots} 
&\cellcolor{gray!15}58.3 \textcolor{mygreen}{\scriptsize$\uparrow$8.3} 
&\cellcolor{gray!15}27.9 \textcolor{gray}{\scriptsize$-$0.0}
&\cellcolor{gray!15}34.5 \textcolor{gray}{\scriptsize$-$0.0} 
&\cellcolor{gray!15}40.0 \textcolor{mygreen}{\scriptsize$\uparrow$6.7} 
&\cellcolor{gray!15}35.4 \textcolor{mygreen}{\scriptsize$\uparrow$2.1} 
&\cellcolor{gray!15}\textbf{35.0 \textcolor{mygreen}{\scriptsize$\uparrow$10.0}} 
&\cellcolor{gray!15}25.0 \textcolor{red}{\scriptsize$\downarrow$3.1} 
&\cellcolor{gray!15}38.5 \textcolor{mygreen}{\scriptsize$\uparrow$1.3} 
&\cellcolor{gray!15}\textcolor{mygreen}{3.16$\uparrow$} \\

\multicolumn{1}{l|}{} & \textbf{15 shots} &33.3 \textcolor{mygreen}{\scriptsize$\uparrow$8.3} &32.6 \textcolor{mygreen}{\scriptsize$\uparrow$2.4} &31.0 \textcolor{mygreen}{\scriptsize$\uparrow$6.9} &31.1 \textcolor{gray}{\scriptsize$-$0.0} &27.1 \textcolor{gray}{\scriptsize$-$0.0} &20.0 \textcolor{gray}{\scriptsize$-$0.0} &37.5 \textcolor{mygreen}{\scriptsize$\uparrow$3.1} &41.0 \textcolor{red}{\scriptsize$\downarrow$1.3} &\textcolor{mygreen}{2.43$\uparrow$} \\
\bottomrule[1.1pt]
\end{tabular}
}
\label{tab2}
\end{table*}

\subsection{Experimental Results}

The results presented in Tables~\ref{tab1} and \ref{tab2} compellingly demonstrate the effectiveness of TaLo. Specifically, for the LLaVA model, TaLo achieves a peak performance gain of \textbf{10.4\%} across five benchmarks and ten different tasks. Notably, its average performance is consistently higher than the baseline model across all three shot settings. Similarly, on the same set of tasks, the Qwen-VL model shows a maximum performance increase of \textbf{16.6\%}. For InternVL, as shown in Table~\ref {tab2}, while our evaluation was conducted on a smaller set of eight tasks, the findings are consistent: TaLo delivers an average performance improvement in all configurations, with a peak gain reaching \textbf{10.0\%}.

\vspace{2em}
\subsection{More Analyses}
\label{analysis}
In this subsection, we provide a streamlined yet comprehensive analysis of our method using MMStar~\citep{chen2024we}, which is a consolidated benchmark spanning six diverse VLM task categories, to validate effectiveness and uncover key patterns systematically.

\vspace{1em}
\noindent{\textbf{Comparison Study. }} To further evaluate TaLo, we compare it with model merging~\citep{chen2025bringreasonvisionunderstanding,yang2024modelmergingllmsmllms,ilharco2023editingmodelstaskarithmetic} as well as fine\mbox{-}tuning methods (LoRA~\citep{hu2021loralowrankadaptationlarge}, OFT~\citep{qiu2024controllingtexttoimagediffusionorthogonal}), all aiming to improve task\mbox{-}specific performance. We emphasize that while fine-tuning can possibly serve as a stronger baseline when abundant labeled data is available, our goal is not to claim superiority over FT in data-rich regimes. Rather, under the same extremely low-data setting used for TaLo we find that parameter-efficient fine-tuning methods like LoRA struggle to learn effectively, often failing to outperform the unmodified base model. For all fine\mbox{-}tuning experiments, we select the checkpoint from the epoch that achieves the best validation performance upon convergence. We use LLaVA\mbox{-}Next\mbox{-}LLaMA3\mbox{-}8B as the base VLM, and the same few-shot samples for both TaLo and the fine\mbox{-}tuning approaches. Detailed configurations are provided in Appendix~\ref{appendix:experimental_details}. As shown in Table~\ref{tab3}, TaLo achieves superior performance compared to both merging and fine\mbox{-}tuning baselines in less time across most settings. Crucially, unlike any of the baselines, TaLo requires neither external models nor any form of training or task\mbox{-}specific parameter updates. Instead, it enables on\mbox{-}the\mbox{-}fly adaptation through minimal, dynamic intervention within the base model, highlighting its efficiency, simplicity, and strong practical applicability.

\vspace{0.35em}
\noindent{\textbf{Extending TaLo to Multi\mbox{-}Layer Interventions. }} 
To assess whether TaLo’s effectiveness generalizes beyond single-layer intervention, we extend the method to jointly intervene on pairs of layers. Specifically, after identifying the best-performing single layer for a given task, we systematically test adding a second intervention across all remaining layers and select the pair that yields the highest performance gain. Our findings, as reported in Appendix\ref{mli} reveal that for many tasks, introducing a second intervention provides little or no improvement over the best single-layer result, and in several cases, no viable second task-interfering layer can be identified at all. This suggests that task-interfering layers are likely sparse and that complex inter-layer dependencies may obscure their individual roles—further justifying TaLo’s targeted single-layer approach. (See Appendix~\ref{more results of TaLo} for additional results and analysis.)

\begin{table}[t!]
\renewcommand{\arraystretch}{1.2}
\caption{Comparison of TaLo with other methods. Each entry reports performance and running time ($\times 10^{2}$ s). TaLo attains higher accuracy with lower adaptation time.} 
\begin{minipage}[t]{0.495\textwidth}
\centering
\resizebox{\linewidth}{!}{%
\begin{tabular}{lcccccccc}
    \toprule[1.2pt]
    \multicolumn{9}{c}{\textbf{Math}} \\
    \midrule
    \multirow{2}{*}{\textbf{Shots}} & \textbf{Base} & \textbf{Merge} & \multicolumn{2}{c}{\textbf{LoRA-FT}} & \multicolumn{2}{c}{\textbf{OFT}} & \multicolumn{2}{c}{\textbf{TaLo}} \\
    \cmidrule(lr){4-5} \cmidrule(lr){6-7} \cmidrule(lr){8-9}
    & \textbf{Score$\uparrow$} & \textbf{Score$\uparrow$} & \textbf{Score$\uparrow$} & \textbf{Time$\downarrow$} & \textbf{Score$\uparrow$} & \textbf{Time$\downarrow$} & \textbf{Score$\uparrow$} & \textbf{Time$\downarrow$} \\
    \midrule
    \textbf{10shots} & 30.42 & 32.38 & 31.25 & 1.75 & 30.83 & 14.67 & \textbf{32.92} & 1.70 \\
    \textbf{15shots} & 30.21 & 32.63 & 32.17 & 2.53 & 32.34 & 21.91 & \textbf{33.19} & 1.76 \\
    \textbf{20shots} & 30.87 & 31.76 & 33.04 & 3.46 & 33.04 & 29.16 & \textbf{33.91} & 2.84 \\
    \midrule
    \textbf{Avg} & 30.50 & 32.26 & 32.15 & 2.58 & 32.07 & 21.91 & \textbf{33.34} & 2.10 \\
    \bottomrule[1.1pt]
\end{tabular}
}
\end{minipage}%
\vspace{1em}
\begin{minipage}[t]{0.495\textwidth}
\centering
\resizebox{\linewidth}{!}{%
\begin{tabular}{lcccccccc}
    \toprule[1.2pt]
    \multicolumn{9}{c}{\textbf{Instance Reasoning}} \\
    \midrule
    \multirow{2}{*}{\textbf{Shots}} & \textbf{Base} & \textbf{Merge} & \multicolumn{2}{c}{\textbf{LoRA-FT}} & \multicolumn{2}{c}{\textbf{OFT}} & \multicolumn{2}{c}{\textbf{TaLo}} \\
    \cmidrule(lr){4-5} \cmidrule(lr){6-7} \cmidrule(lr){8-9}
    & \textbf{Score$\uparrow$} & \textbf{Score$\uparrow$} & \textbf{Score$\uparrow$} & \textbf{Time$\downarrow$} & \textbf{Score$\uparrow$} & \textbf{Time$\downarrow$} & \textbf{Score$\uparrow$} & \textbf{Time$\downarrow$} \\
    \midrule
    \textbf{10 shots} & 54.16 & 53.33 & 55.83 & 2.06 & 55.42 & 15.34 & \textbf{58.33} & 1.33 \\
    \textbf{15 shots} & 53.61 & 52.34 & 55.32 & 2.87 & 56.17 & 25.43 & \textbf{57.87} & 1.81 \\
    \textbf{20 shots} & 54.35 & 53.04 & 55.65 & 4.06 & 56.52 & 33.81 & \textbf{58.69} & 2.89 \\
    \midrule
    \textbf{Avg} & 54.04 & 52.90 & 55.60 & 2.99 & 56.04 & 24.86 & \textbf{58.30} & 2.01 \\
    \bottomrule[1.1pt]
\end{tabular}
}
\end{minipage}
\label{tab3}
\end{table}

% Due to the rapid growth in computational cost when exploring multi\mbox{-}layer interventions, we conduct our analysis using two\mbox{-}layer combinations as a representative sample. Despite this simplification, our findings remain highly informative. As shown in Table~\ref{tab:2layers}, for several tasks, no second task\mbox{-}interfering layer can be identified, suggesting task\mbox{-}interfering layers may be sparse and strong inter\mbox{-}layer interactions further obscure individual roles, thereby justifying TaLo’s focus on single\mbox{-}layer interventions.

\section{Discussions}
\label{disscussion}
\noindent{\textbf{Limitation and Future Work.}} Our analysis relies on existing benchmarks, and the predefined task categories within them may influence the layer sensitivity patterns we observed. Future work could therefore validate our findings across more granular task decompositions, which may help establish the generality of this phenomenon. With respect to our method, TaLo, we acknowledge that it is a minimalist framework. Our primary objective was not to achieve state\mbox{-}of\mbox{-}the\mbox{-}art performance, but rather to provide a simple, plug\mbox{-}and\mbox{-}play solution that serves as a proof\mbox{-}of\mbox{-}concept for the practical utility of the Task\mbox{-}Interfering Layer phenomenon. We also explored extensions, such as the multi-layer interventions, which showed that identifying multiple interfering layers is non-trivial. This finding supports our focus on a minimalist, single-layer approach as a robust and practical starting point. There are several promising avenues to explore, such as developing more sophisticated dynamic layer selection mechanisms, investigating better multi\mbox{-}layer modulation strategies, or more fine-grained intervention operations. Furthermore, this layer-wise diagnostic signal holds potential for guiding model training; for instance, by applying targeted regularization to the identified interfering layers, one might mitigate task conflicts at their source. We are optimistic that future research can extend TaLo's capabilities, applying its principles to a broader and more complex range of scenarios.

\noindent{\textbf{Hypothesis and Explanation.}} We further offer a hypothesis to explain why certain layers may become task\mbox{-}interfering. Modern large models are pretrained on diverse and multi\mbox{-}task data, where each layer learns a compromise representation, which approximates a global optimum across all tasks. However, this global optimum may deviate from the local optimum for any specific task. We conjecture that Task\mbox{-}Interfering Layers capture features that, while beneficial on average, introduce noise or misalignment when applied to a particular task. By zeroing out or uniformly scaling these layers, TaLo effectively suppresses or rebalances their influence, which may prevent the propagation of task\mbox{-}irrelevant or even detrimental information. This intervention, we hypothesize, steers the model’s internal computation toward a more favorable region in the parameter space that better aligns with the target task’s local optimum, thereby  effectively improving performance without any parameter updates.

\section{Conclusion}
{
\hyphenpenalty=10000
\tolerance=5000
Through extensive empirical analysis, we reveal the existence of specific layers within large-scale pretrained Vision-Language models that actively suppress performance on certain downstream tasks. We term these Task-Interfering Layers, as strategically bypassing them yields significant performance improvements. Our further investigation uncovers a crucial pattern: tasks that demand similar functional abilities exhibit highly consistent response patterns to layer interventions. This suggests that the interference phenomenon is systematically organized around the model's functional capabilities, allowing the effects of Task-Interfering Layers to generalize across related tasks. Based on these findings, we introduce TaLo, a training-free adaptation method that identifies and bypasses these interfering layers at inference time. The performance of TaLo across a diverse range of models demonstrates that simple, targeted layer intervention can be a efficient strategy for model adaptation, obviating any parameter updates.
}

{
    \small
    \bibliographystyle{ieeenat_fullname}
    \bibliography{main}
}

\appendix
\clearpage
\setcounter{page}{1}
\maketitlesupplementary

\section{Models and Benchmarks}
\label{Models and Benchmarks}
We present all the models used in our experiments in Table \ref{tab:model_details}, and list all the benchmarks we utilize in Table \ref{tab:bench_details}. 

\begin{table*}
\centering
\scalebox{1}{
\begin{tabular}{l l p{7cm}}
\toprule

Name & Size & Huggingface ckpt \\
\midrule
LLaVA-Next-LLaMA3 \citep{li2024llavanext-strong} & 8B & \url{llava-hf/llama3-llava-next-8b-hf} \\
Qwen2-VL \citep{wang2024qwen2} & 2B & \url{Qwen/Qwen2-VL-2B-Instruct} \\
InternVL2 \citep{chen2024internvl} & 26B \& 40B & \url{OpenGVLab/InternVL2-40B} \\
\bottomrule
\end{tabular}
}
\caption{Details of the models used in our experiments.}
\label{tab:model_details}
\end{table*}

\begin{table*}
\centering
\begin{tabular}{l l p{8cm}}
\toprule
Benchmark & Category & Huggingface URL \\
\midrule
MMStar \citep{chen2024we} & MCQ & \url{Lin-Chen/MMStar} \\
MMBench-EN \citep{liu2024mmbench} & MCQ & \url{lmms-lab/MMBench} \\
MMMU-VAL \citep{yue2023mmmu} & MCQ & \url{MMMU/MMMU} \\
ScienceQA-VAL \citep{lu2022learn} & MCQ & \url{derek-thomas/ScienceQA} \\
MathVista-MINI \citep{lu2024mathvista} & VQA & \url{AI4Math/MathVista} \\
SEEDBench-IMG \citep{li2023seed} & MCQ & \url{lmms-lab/SEED-Bench} \\
\bottomrule
\end{tabular}
\caption{Details of the benchmarks used in our experiments.}
\label{tab:bench_details}
\end{table*}
\vspace{0.5em}
In this work, all datasets are evaluated using accuracy as the sole metric. The majority of datasets: MMStar, MMMU, SEEDBench, MMBench, and ScienceQA are multiple-choice question (MCQ) benchmarks, where the model’s predicted option is extracted from its output and matched against the ground truth. MathVista, while formulated as a vision-question-answering (VQA) task, also employs direct string matching between generated responses and reference answers in its official evaluation, ensuring consistency in the metric across all tasks.

Specifically, MMStar is a comprehensive benchmark with 250 balanced samples across six core capabilities: Coarse Perception, Fine-grained Perception, Instance Reasoning, Logical Reasoning, Math, and Science \& Technology. MMBench contains 2,974 MCQs assessing a wide range of abilities, including Coarse Perception, Fine-grained Perception (both single and cross instance), Instance Reasoning, Logic Reasoning, Attribute Reasoning, and Relation Reasoning. MMMU spans 30 disciplines, including Art \& Design, Business, Science, Health \& Medicine, Humanities \& Social Sciences, and Engineering, covering 183 subfields with 30 types of heterogeneous images (e.g., charts, diagrams, maps, tables, musical scores, chemical structures), focusing on advanced perception and reasoning with domain-specific knowledge. SEEDBench comprises 19,000 human-annotated MCQs, covering 12 evaluation dimensions, including image understanding. MathVista is a challenging benchmark that combines diverse mathematical and visual reasoning tasks, consisting of 6,141 examples drawn from 28 existing multimodal math-related datasets and three newly curated datasets. Finally, ScienceQA consists of 21,208 multimodal science questions collected from elementary and high school curricula. This diverse and rigorous selection of benchmarks enables a comprehensive evaluation of task-specific abilities under a unified accuracy metric.

\section{Experimental Details}
\label{appendix:experimental_details}
In our TaLo experiments, the procedure for identifying the optimal intervention layer begins with task definition and sample preparation. We first identify the target task according to the dataset's metadata, after which we draw samples from a probing pool held entirely separate from the final test set to prevent any data overlap.

Our identification process follows an iterative pipeline. We first establish a baseline performance by evaluating the unmodified model on an initial set of probing samples. If the baseline accuracy reaches 100\%, the sample set is considered uninformative and is discarded; a new set is then drawn from the probing pool, and the baseline is re-evaluated. Once the baseline is established, we proceed with a systematic, layer-by-layer parameter intervention and measure the performance gain for each. If a unique layer yields the maximum positive gain, it is designated as the optimal target. In cases where multiple layers tie for the best performance or no layer produces a positive gain, we initiate a multi-round, augmented sampling strategy to resolve the ambiguity. This involves supplementing the set with an additional $shot/2$ samples for re-evaluation, followed by a further $shot/4$ samples if the tie persists. Should a unique optimal layer still not be identified after these two rounds, we select the layer with the highest index among the final candidates to ensure robustness\citep{yin2023outlier,gromov2024unreasonable,sun2025curse,men2024shortgpt}.

For our fine-tuning experiments, we employ two parameter-efficient fine-tuning (PEFT) methods: LoRA (Low-Rank Adaptation) and its variant OFT (Orthogonal Finetuning). All experiments are conducted on the LLaVA-Next-8B model. In the case of LoRA, we set the rank $r=8$ and scaling factor $\alpha=16$, and apply the adapter modules to all linear projections in both the language and vision pathways. This full-architecture adaptation strategy ensures comprehensive alignment of both visual and textual representations during fine-tuning. To ensure that the model truly understands the knowledge underlying the questions during fine-tuning, rather than simply memorizing the options, we format the answers as ``option + option content''. This approach helps the model learn the specific meaning of each option and its relationship to the question.

For model merging experiments, the LLM used for merging is DeepSeek-R1-Distill-Llama-8B\citep{deepseekai2025deepseekr1incentivizingreasoningcapability}, with a fusion coefficient $\lambda$ of 0.9. However, as shown in Table~\ref{tab3}, the effectiveness of model merging is highly sensitive to both the choice of the external LLM and the target task, suggesting that its performance is not robust across configurations and requires careful, task-specific tuning.
\section{Additional Experimental Results and\\ Analysis}
\label{more results}

\subsection{Specific clustering details}
\begin{figure}[t!]
    \centering
    \caption{Qualitative case study on random noise intervention.}
    \includegraphics[width=1\linewidth]{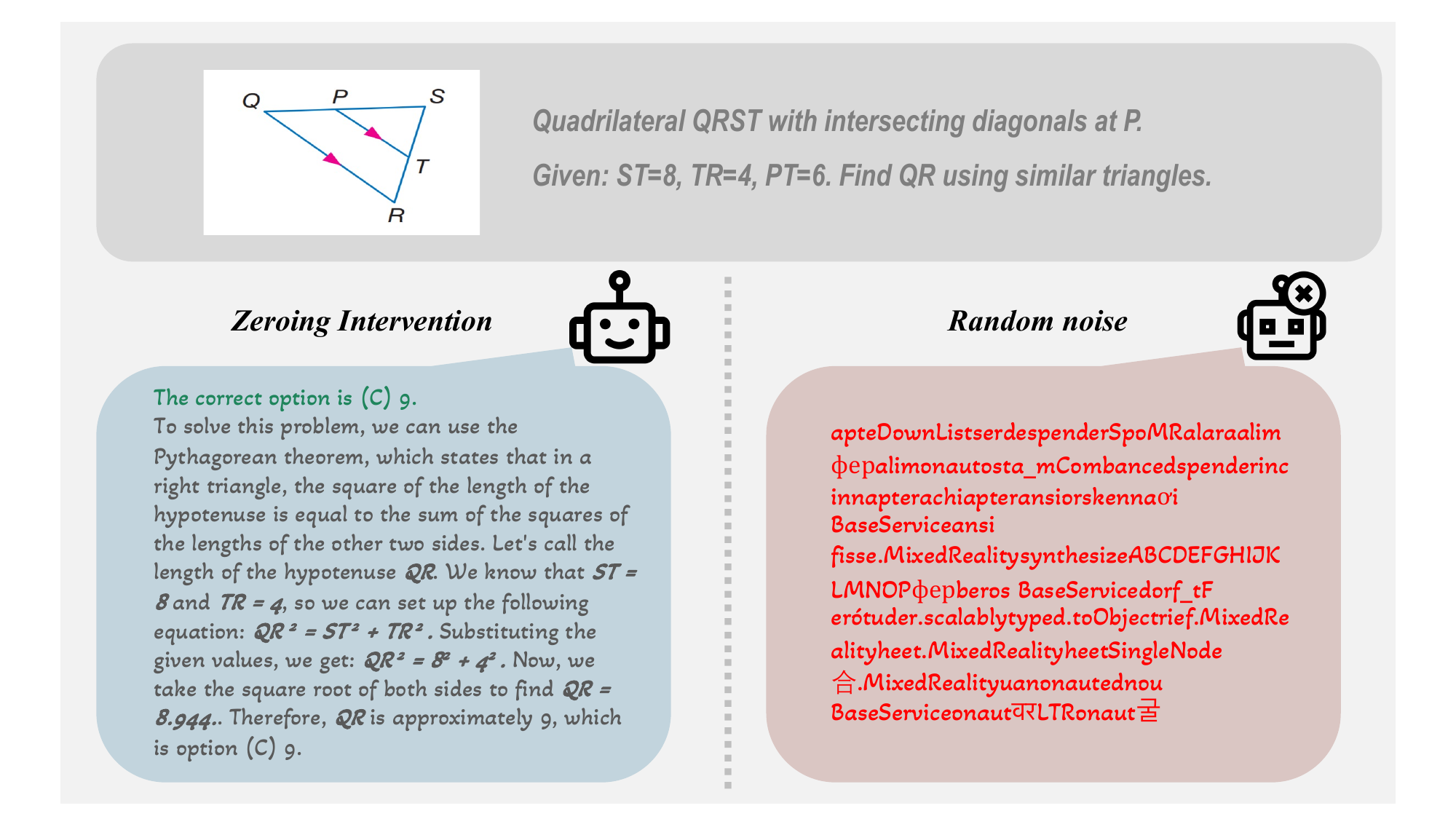}
    \label{fig:random}
\end{figure}
Table~\ref{tab:clusters_tasks} provides the comprehensive list of tasks included in each of the seven clusters. From the table, the clustering appears to meaningfully group tasks by functional similarity. For instance, Cluster 1 brings together \texttt{numeric commonsense}, \texttt{arithmetic reasoning}, and \texttt{math word problem}—all clearly centered on numerical understanding and calculation. This suggests the method successfully identifies and isolates quantitative reasoning as a coherent capability.

Similarly, Cluster 3 stands out by grouping domain-specific scientific tasks—\texttt{Astronomy}, \texttt{Chemistry}, and \texttt{Scientific Reasoning}—into a unified theme, reflecting shared reliance on formal scientific knowledge.

\subsection{Ablation on Intervention Component}
\label{ablation on ffn}
In the main paper, we defines layer intervention as the modification of self-attention modules. To justify this choice, we conducted comprehensive ablation experiments applying the same intervention to the MLP modules.

\begin{figure}[t]
  \caption{Accuracy change after intervening on MLP layers}
  \centering
  % 第1行
  \begin{subfigure}[b]{0.48\columnwidth}
    \centering
    \includegraphics[width=\linewidth]{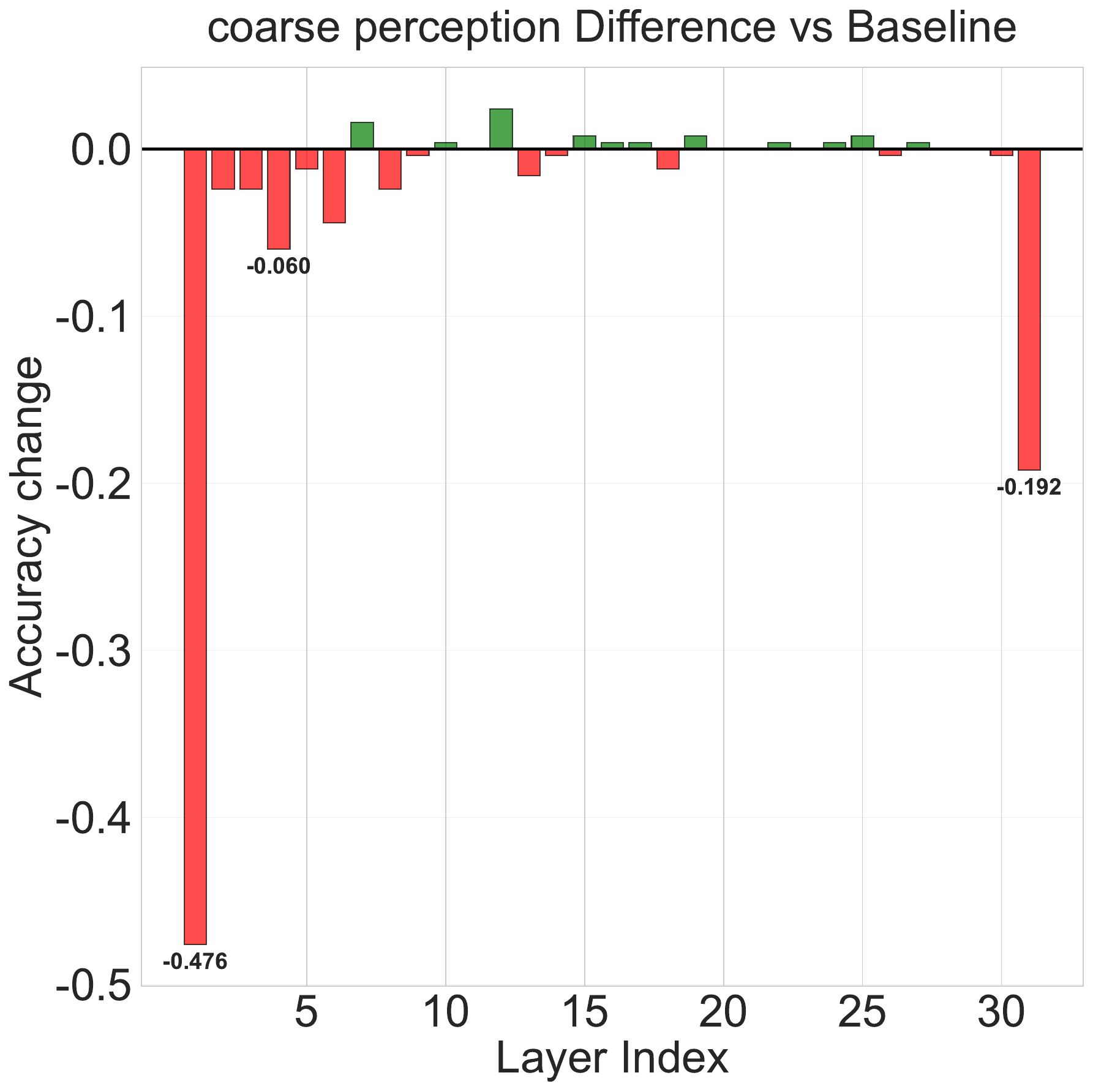}
  \end{subfigure}\hfill
  \begin{subfigure}[b]{0.48\columnwidth}
    \centering
    \includegraphics[width=\linewidth]{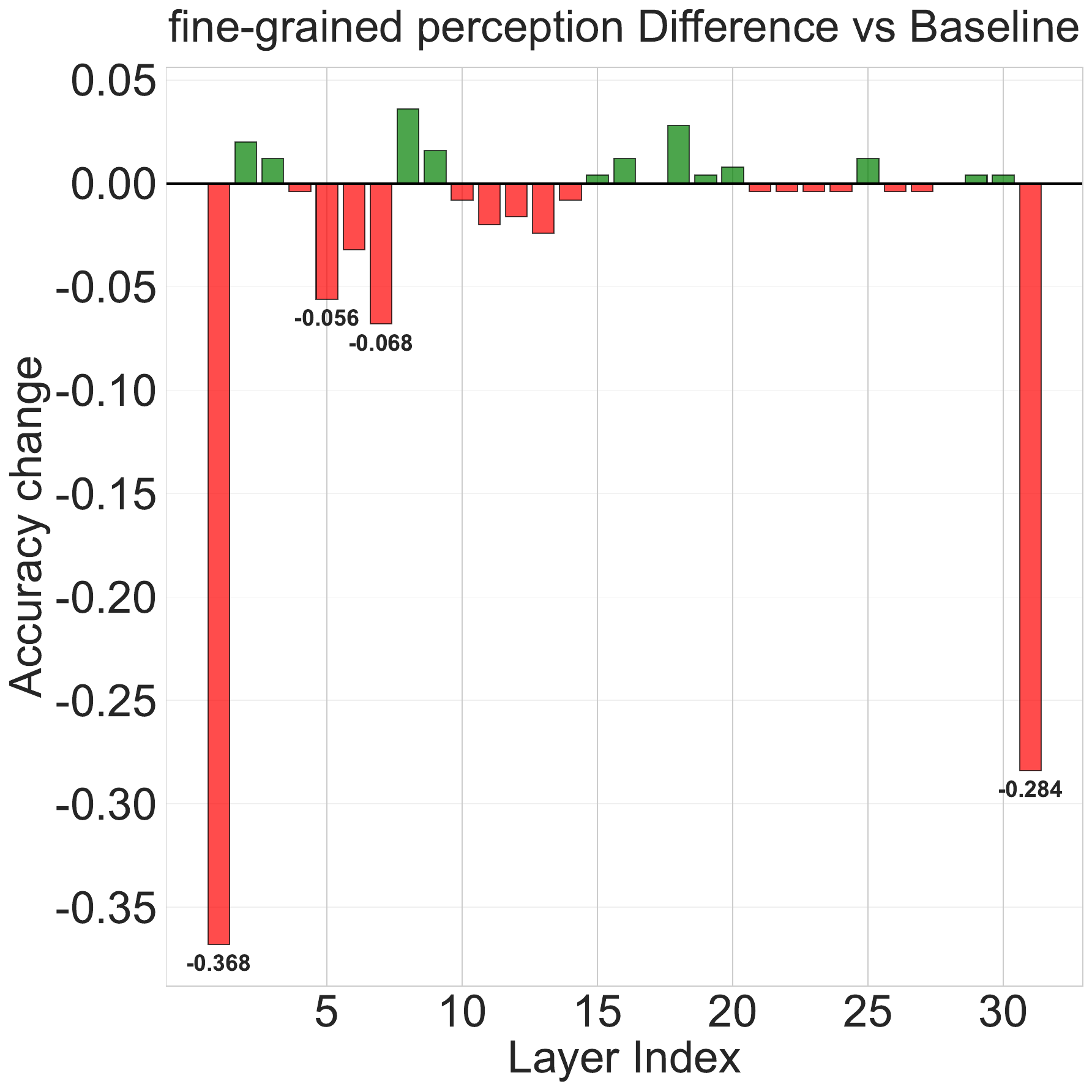}
  \end{subfigure}

  \vspace{6pt}
  % 第2行
  \begin{subfigure}[b]{0.48\columnwidth}
    \centering
    \includegraphics[width=\linewidth]{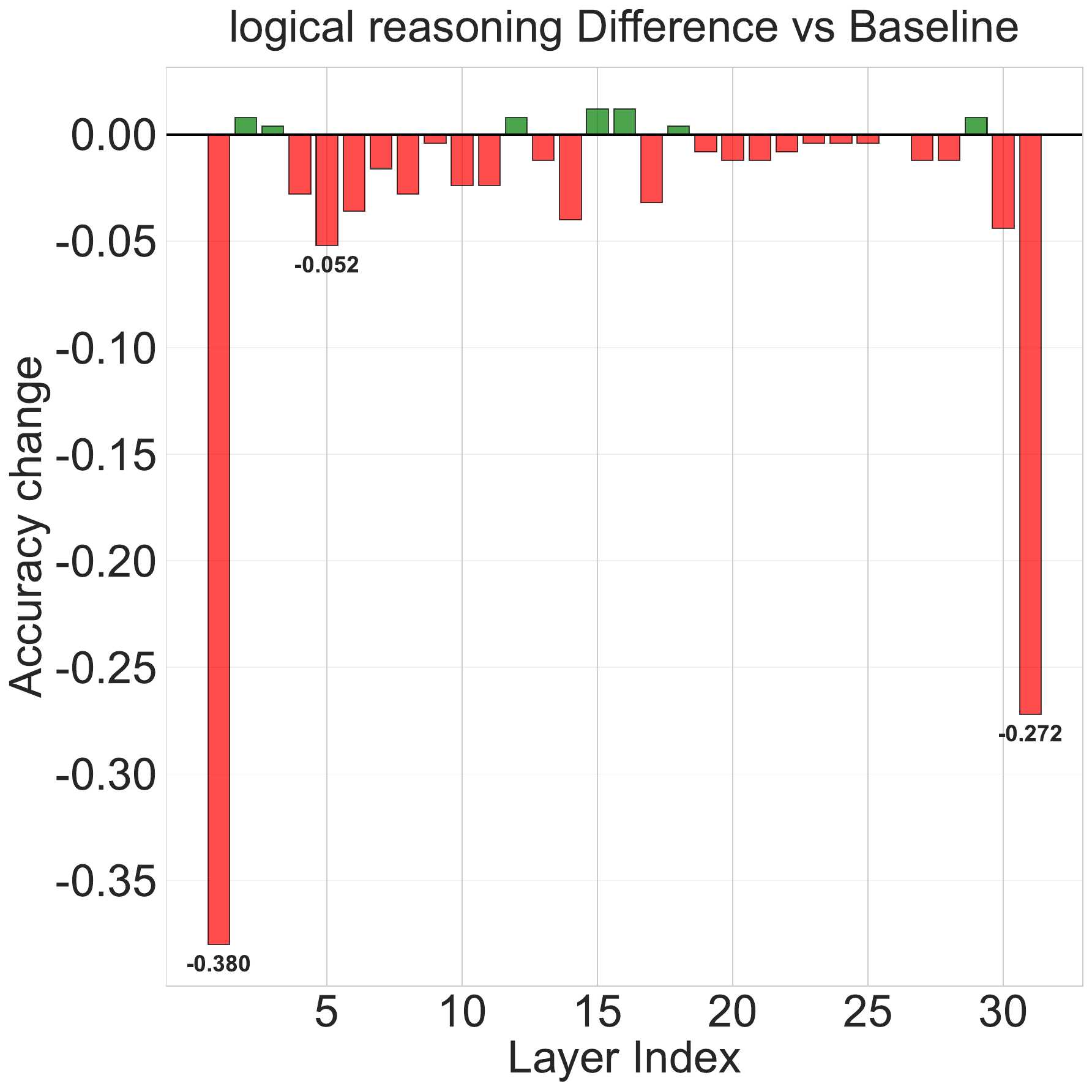}
  \end{subfigure}\hfill
  \begin{subfigure}[b]{0.48\columnwidth}
    \centering
    \includegraphics[width=\linewidth]{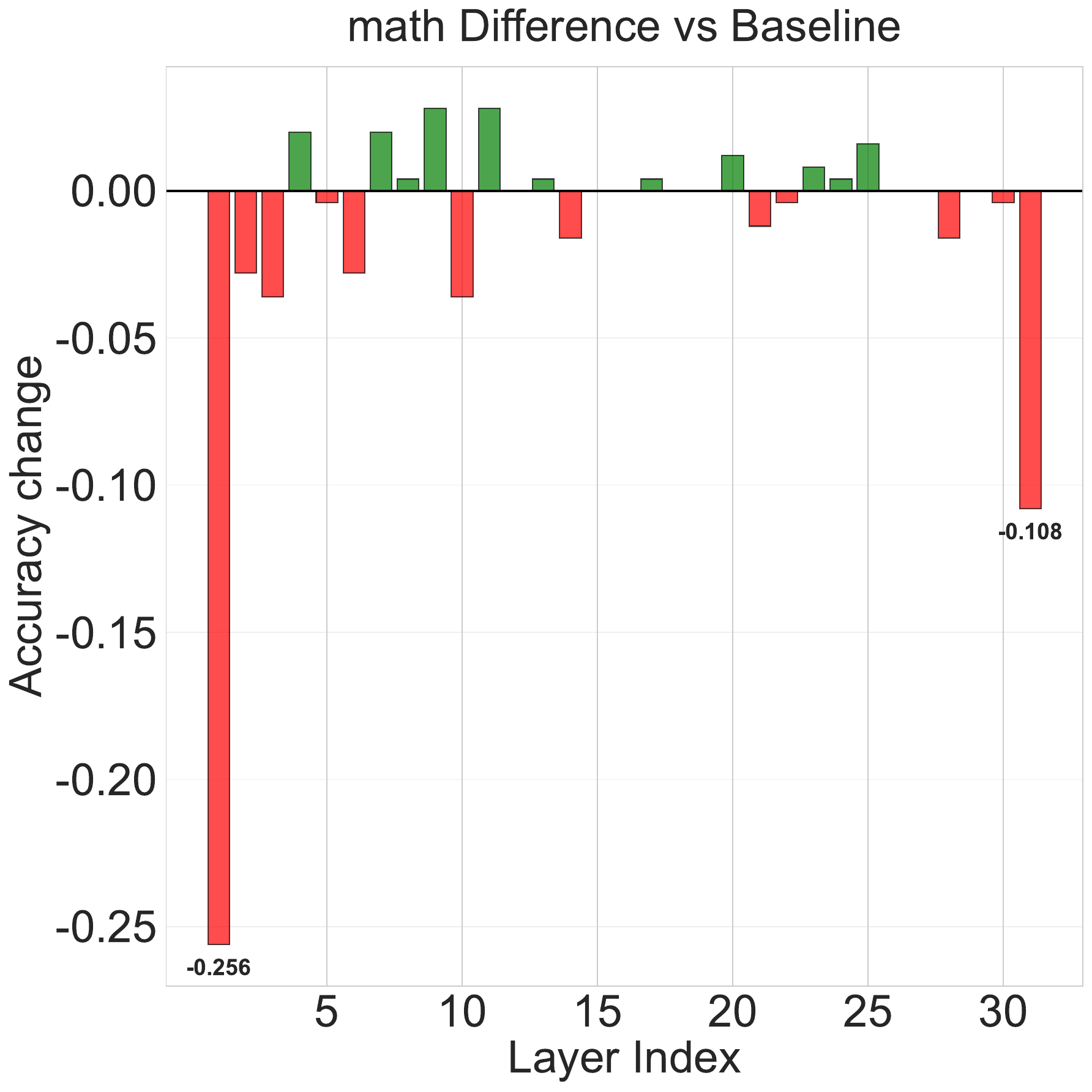}
  \end{subfigure}

  \vspace{6pt}
  % 第3行
  \begin{subfigure}[b]{0.48\columnwidth}
    \centering
    \includegraphics[width=\linewidth]{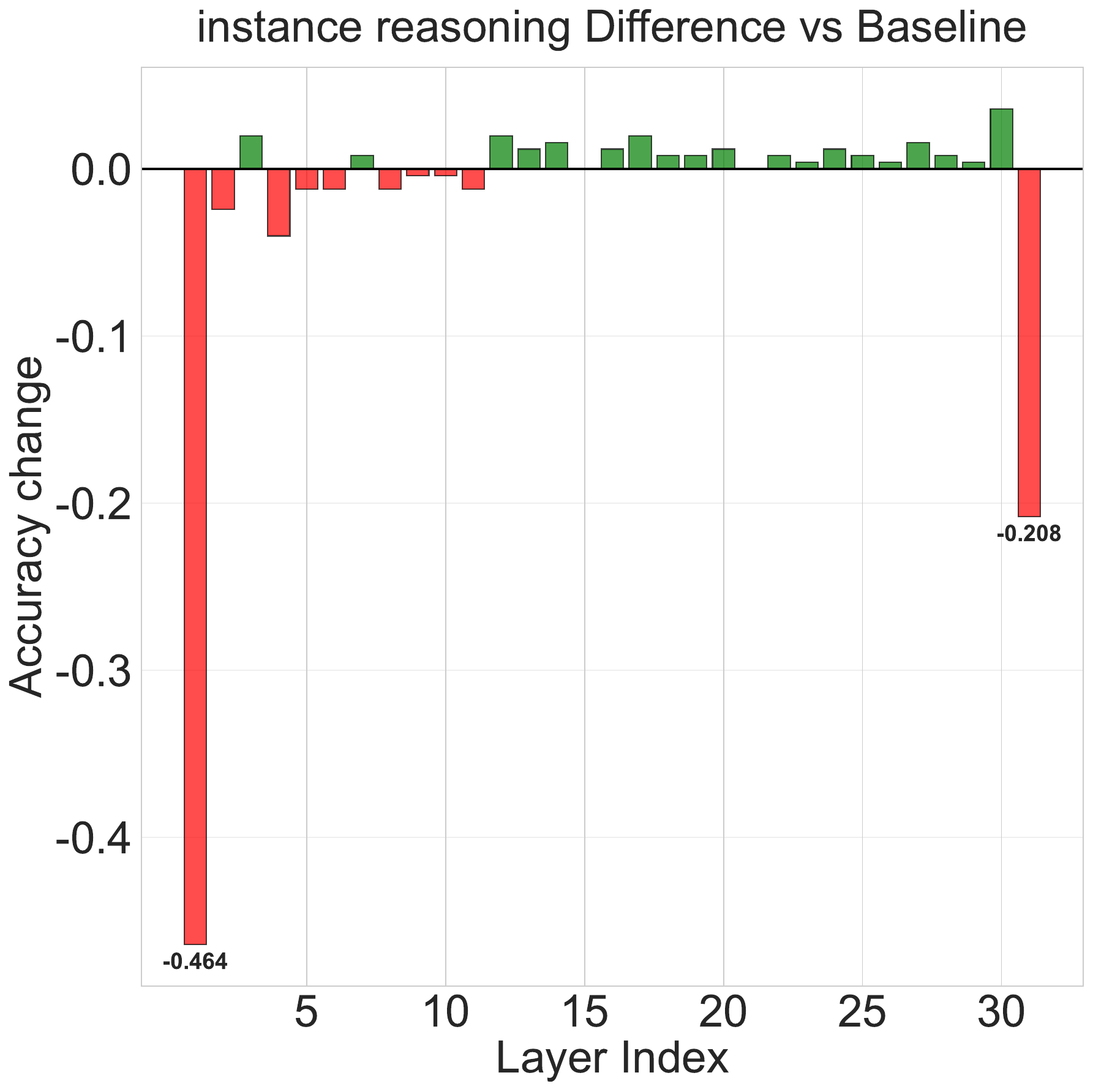}
  \end{subfigure}\hfill
  \begin{subfigure}[b]{0.48\columnwidth}
    \centering
    \includegraphics[width=\linewidth]{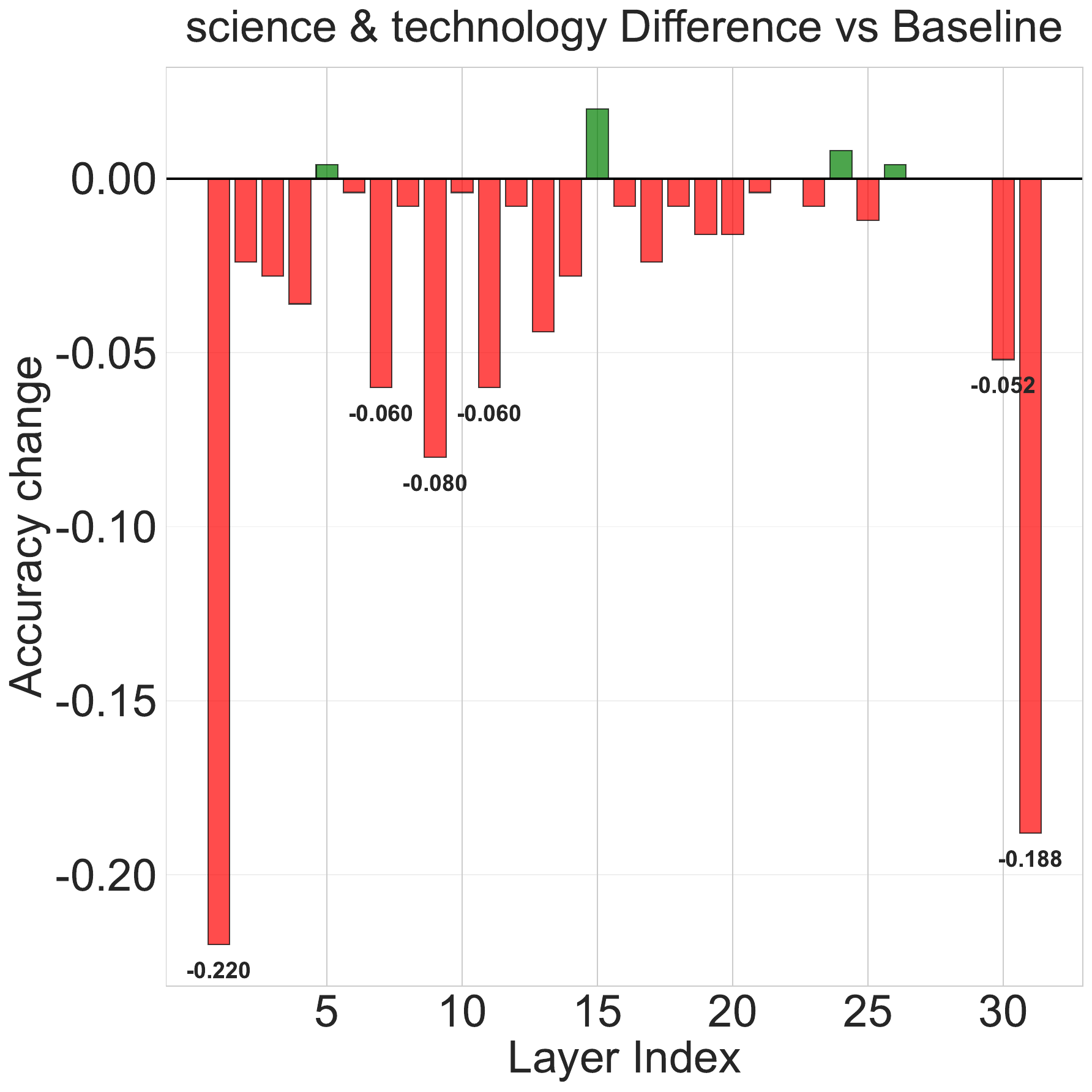}
  \end{subfigure}
  \label{fig:2x3grid}
\end{figure}

As shown in Figure~\ref{fig:2x3grid}, the intervention effect on MLP modules demonstrates a strong position-dependent sensitivity and a pattern fundamentally different from that of self-attention modules. We found that intervening on the early layers or the final layer of the model results in a catastrophic performance collapse across all tasks. As for the vast majority of middle layers, while intervening on the MLP module does not cause model collapse, showing a degree of robustness, it also fails to produce any significant performance gains. This finding justifies our decision to focus this study on self-attention modules, as they clearly exhibit the significant performance-boosting Task-Interfering Layer phenomenon we aim to investigate. Consequently, all subsequent analysis and the proposed TaLo method operate exclusively on these self-attention modules.

\subsection{Additional Analysis of TaLo}

\paragraph{Ablation study of the Intervention Method}
While earlier results (Section ~\ref{consistency}) suggest that different intervention types induce similar layer-wise trends overall, their impact can vary subtly across individual tasks. To examine these influences in a controlled yet representative setting, we turn to MMStar and MMBench: two balanced benchmarks that aggregate data from multiple sources, offering moderate task diversity and comprehensive coverage. It avoids the extremes of highly specialized or overly narrow datasets, making it well-suited for studying how these factors interact with TaLo’s behavior.

In addition to parameter zeroing and uniform scaling, we also explored replacing layer weights with their mean value and random noise. However, experiments show that injecting random noise severely destabilizes the model, effectively erasing the learned representations (as shown in Figure~\ref{fig:random}). The outputs become incoherent, often degenerate, with little connection to the input. This suggests that the pre-trained parameters, even when suboptimal for a specific task, still encode essential structural and semantic priors critical to model functionality.

Given this breakdown in basic competence, we focus our analysis on milder, more controlled interventions: zeroing, uniform scaling, and mean replacement, which preserve the model’s foundational knowledge while allowing targeted modulation. These methods strike a better balance between perturbation and stability, enabling meaningful study of layer-wise task adaptation without collapsing overall performance.

\begin{table*}[htbp]
\centering
\renewcommand{\arraystretch}{1.3} % 调整行距
\caption{Results of TaLo on \textit{LLaVA-Next-Llama3-8B} under different intervention methods. The last column (\textbf{Avg}) reports the mean change across tasks. \ding{55} indicates the method failed to find the Task\mbox{-}Interfering layer.}
\scalebox{0.75}{
\begin{tabular}{lcccccc|c}
\toprule[1.2pt]
\multirow{2}{*}{\textbf{Intervention}} 
& \multicolumn{6}{c|}{\textbf{MMStar}} & \multirow{2}{*}{\textbf{Avg}} \\ 
\cmidrule(lr){2-7}
& \textbf{Coarse perception} & \textbf{Fine-grained perception} & \textbf{Instance reasoning} & \textbf{Science \& technology} & \textbf{Logical reasoning} & \textbf{Math} & \\ 
\midrule
\textbf{Zeroing} 
& 63.8 \textcolor{red}{\scriptsize$\downarrow$2.9} 
& 41.9 \textcolor{mygreen}{\scriptsize$\uparrow$0.9} 
& 57.6 \textcolor{mygreen}{\scriptsize$\uparrow$3.8} 
& 32.9 \textcolor{mygreen}{\scriptsize$\uparrow$1.0} 
& 38.6 \textcolor{red}{\scriptsize$\downarrow$3.8} 
& 31.0 \textcolor{mygreen}{\scriptsize$\uparrow$2.9} 
& \textcolor{mygreen}{0.32$\uparrow$} \\

\textbf{Uniform scaling} 
& 66.2 \textcolor{mygreen}{\scriptsize$\uparrow$2.4} 
& 41.0 \textcolor{red}{\scriptsize$\downarrow$0.4} 
& 51.0 \textcolor{red}{\scriptsize$\downarrow$2.8} 
& 24.8 \textcolor{red}{\scriptsize$\downarrow$5.2} 
& 40.0 \textcolor{red}{\scriptsize$\downarrow$1.0} 
& 34.3 \textcolor{mygreen}{\scriptsize$\uparrow$3.8} 
& \textcolor{red}{0.53$\downarrow$} \\

\textbf{Mean replacement} 
& {\ding{55}} 
& 34.8 \textcolor{red}{\scriptsize$\downarrow$6.2}  
& 52.9 \textcolor{red}{\scriptsize$\downarrow$0.9}  
& 27.6 \textcolor{red}{\scriptsize$\downarrow$3.4}  
& 41.9 \textcolor{mygreen}{\scriptsize$\uparrow$1.4} 
& 30.5 \textcolor{mygreen}{\scriptsize$\uparrow$1.0} 
& \textcolor{red}{1.14$\downarrow$} \\
\bottomrule[1.1pt]
\end{tabular}
}

\label{tab:mean}
\end{table*}

To further examine the impact of three intervention types, we conduct ablation studies across a wide range of tasks on two benchmark datasets, using a consistent 10-shot setting. As shown in Table~\ref{tab:mean},\ref{tab:uniform}, we observe that zeroing and uniform scaling yield comparable effects, with zeroing achieving better average performance across tasks. In contrast, mean replacement consistently leads to inferior results.

This observation aligns with our earlier findings in Section~\ref{consistency}, where both scaling and zeroing exhibited similar layer sensitivity patterns. While subtle differences may arise in specific contexts, which depend on model architecture or task nature.

In practice, the choice between scaling and zeroing can depend on task-specific behavior or implementation simplicity. Both support effective plug-and-play adaptation. TaLo’s strength appears to lie not in the intervention itself, but in the strategic selection of where and when to apply it. It is the layer not the operation seems to be the more decisive factor.

\begin{table*}
\centering
\renewcommand{\arraystretch}{1.2}
\caption{Results of TaLo on \textit{LLaVA-Next-LLaMA3-8B} under different intervention methods (\textit{Structuralized i-t understanding} stands for Structuralized image text understanding).}
\scalebox{0.75}{
\begin{tabular}{lccccc|c}
\toprule[1.3pt]
\multirow{2}{*}{\textbf{Intervention}} 
& \multicolumn{5}{c|}{\textbf{MMBench}} 
& \multirow{2}{*}{\textbf{Avg}} \\ 
\cmidrule(lr){2-6}
& \textbf{Physical property reasoning} 
& \textbf{Structuralized i-t understanding} 
& \textbf{Attribute recognition} 
& \textbf{Celebrity recognition} 
& \textbf{Image emotion} 
& \\ 
\midrule
\textbf{Zeroing} 
& 55.3 \textcolor{mygreen}{\scriptsize$\uparrow$7.8} 
& 55.8 \textcolor{mygreen}{\scriptsize$\uparrow$0.8} 
& 65.2 \textcolor{red}{\scriptsize$\downarrow$3.1} 
& 68.5 \textcolor{red}{\scriptsize$\downarrow$1.4} 
& 65.6 \textcolor{mygreen}{\scriptsize$\uparrow$0.6} 
& \textcolor{mygreen}{0.94$\uparrow$} \\

\textbf{Uniform Scaling} 
& 57.0 \textcolor{mygreen}{\scriptsize$\uparrow$5.0} 
& 55.8 \textcolor{mygreen}{\scriptsize$\uparrow$0.8} 
& 71.0 \textcolor{mygreen}{\scriptsize$\uparrow$0.5} 
& 67.7 \textcolor{red}{\scriptsize$\downarrow$2.8} 
& 65.6 \textcolor{red}{\scriptsize$\downarrow$0.6} 
& \textcolor{mygreen}{0.58$\uparrow$} \\
\bottomrule[1.3pt]
\end{tabular}
}

\label{tab:uniform}
\end{table*}

\paragraph{Multi\mbox{-}Layer Interventions}
\label{mli}
To complement the main results based on single-layer intervention, we conduct a systematic study of two-layer TaLo interventions. For each task, we first identify the optimal single layer using the standard TaLo procedure. We then fix this layer and iteratively apply a second zeroing intervention to every other layer in the LLM backbone, measuring the resulting performance change while keeping all other components unchanged. This yields a full pairwise intervention matrix for each task, from which we select the top-performing two-layer combination.

Due to the quadratic growth in computational cost with model depth, we limit our exploration to two-layer combinations as a tractable proxy for higher-order interactions. The results consistently show limited or no gain from adding a second intervention, reinforcing the sparsity of task-interfering layers observed in the main paper.
\begin{table*}[htbp] 
\centering 
\renewcommand{\arraystretch}{1.1} 
\small
\caption{Results of TaLo on \textit{LLaVA} under two\mbox{-}layer intervention (10\mbox{-}shot). `\ding{55}' marks cases where a second Task\mbox{-}Interfering Layer could not be identified. Details of the MMStar are provided in Appendix~\ref{Models and Benchmarks}.} 
\setlength{\tabcolsep}{10pt}
\scalebox{1}{ 
\begin{tabular}{lcccccc} 
\toprule[1.2pt] 
\multirow{2}{*}{\textbf{Metric}} & \multicolumn{6}{c}{\textbf{MMStar}} \\ 
\cmidrule(lr){2-7} 
% 使用 \makecell 命令对长表头进行换行
& \textbf{CP} 
& \textbf{FP} 
& \textbf{IR}
& \textbf{S\&T}
& \textbf{LR}
& \textbf{Math} \\ 
\midrule 
\textbf{Task-Interfering Layer} & L1, L6 & L15, L29 & L11, \ding{55} & L31, \ding{55} & L1, L8  & L6, L13 \\ 
\textbf{Performance (two layers)} & 61.9 \textcolor{red}{\scriptsize$\downarrow$3.8} & 40.0 \textcolor{red}{\scriptsize$\downarrow$0.5} & 56.2 \textcolor{mygreen}{\scriptsize$\uparrow$1.0} & 31.0 \textcolor{mygreen}{\scriptsize$\uparrow$0.5} & 37.1 \textcolor{red}{\scriptsize$\downarrow$3.8} & 25.7 \textcolor{red}{\scriptsize$\downarrow$3.8} \\
\midrule
\textbf{Performance (single layer)} & 63.8 \textcolor{red}{\scriptsize$\downarrow$2.9} & 41.9 \textcolor{mygreen}{\scriptsize$\uparrow$0.9} & 57.6 \textcolor{mygreen}{\scriptsize$\uparrow$3.8} & 38.6 \textcolor{red}{\scriptsize$\downarrow$3.8} & 31.0 \textcolor{mygreen}{\scriptsize$\downarrow$2.9} & 32.9 \textcolor{mygreen}{\scriptsize$\uparrow$2.5} \\ 
\bottomrule[1.1pt] 
\end{tabular} 
}
\label{tab:2layers} 
\end{table*}

\begin{figure}[t]
\caption{Layer Selection's Robustness Analysis.}
    \centering
    % 第一张图
    \begin{subfigure}{0.95\linewidth}
        \centering
        \includegraphics[width=\linewidth]{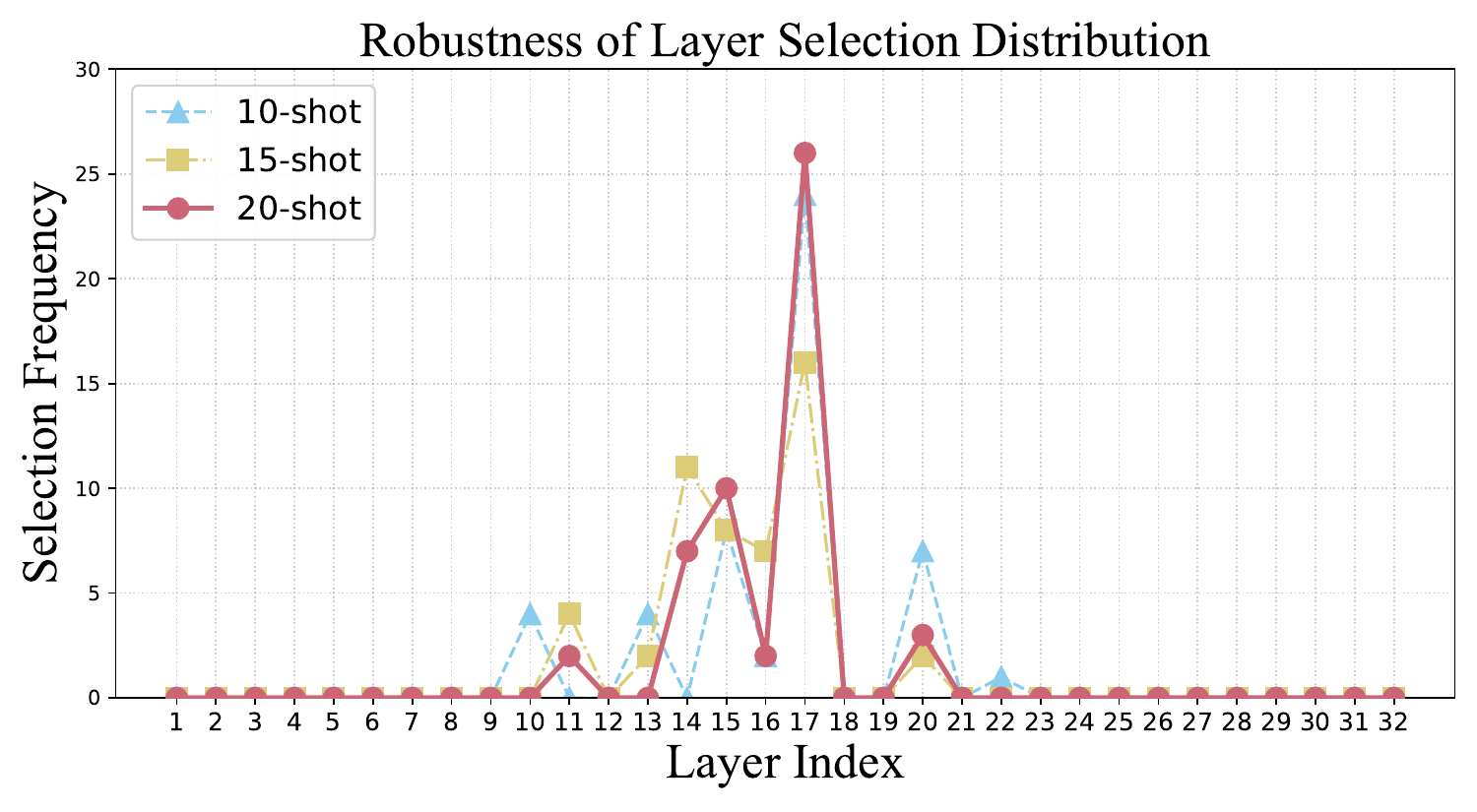}
        \caption{Analysis on Math task}
        \label{fig:top}
    \end{subfigure}
    \vspace{0.5em}
    % 第二张图
    \begin{subfigure}{0.95\linewidth}
        \centering
        \includegraphics[width=\linewidth]{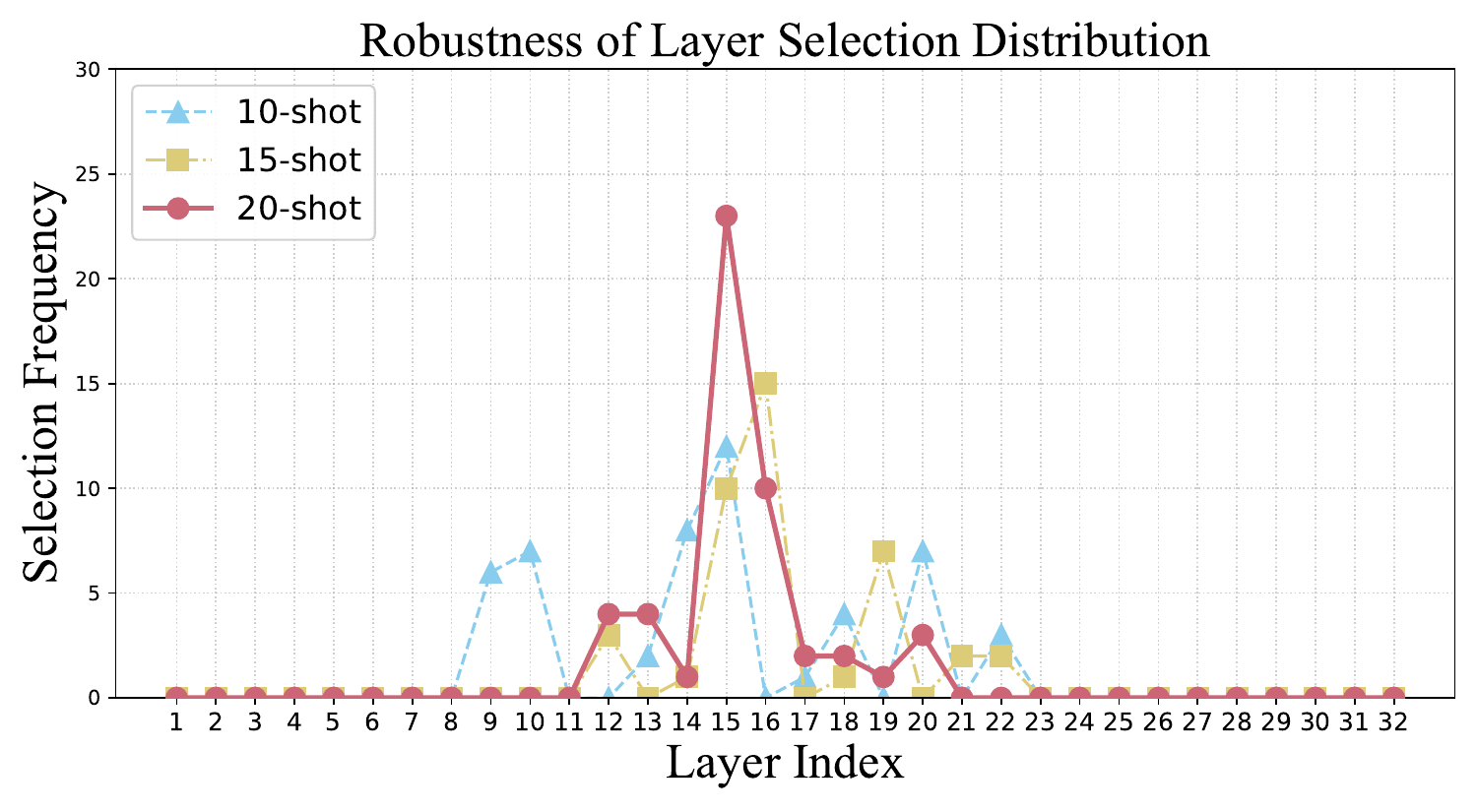}
        \caption{Analysis on Instance-reasoning task}
        \label{fig:bottom}
    \end{subfigure}
    \label{fig:robu}
\end{figure}

\paragraph{Robustness on layer selection}
To demonstrate that the robustness of our layer selection is not limited to a specific domain, we extended the bootstrapping analysis (N=50) to two distinct task categories across three few-shot settings. As illustrated in Figure~\ref{fig:robu}, we observe that the layer selection distribution remains highly concentrated across all tasks, confirming that TaLo consistently localizes a stable region of interference regardless of the task type or sampling variations. Crucially, the distinct locations of these interference regions across different tasks powerfully validate that TaLo captures meaningful, task-specific functional conflicts inherent to the model's internal representations, rather than merely identifying universally redundant layers.

\paragraph{More Experimental Results of TaLo}

\label{more results of TaLo}
As evidenced by the comprehensive results in Table~\ref{tab:llava} \ref{tab:qwen}, which encompasses evaluations on MMBench and ScienceQA, TaLo demonstrates a robust ability to enhance performance across a wide spectrum of tasks. Nonetheless, the magnitude of improvement is observed to be more constrained and in some cases even decreases for particularly challenging categories, such as those involving complex multi-step reasoning, detailed visual attribute discrimination. This pattern suggests that while our proposed layer-level intervention provides an effective mechanism for task adaptation, its efficacy is bounded by the underlying capabilities of the pre-trained model. Performance plateaus or regressions in these demanding scenarios likely point to limitations that are architectural or data-based in nature, which might be addressed in future work by integrating stronger inductive biases or auxiliary knowledge sources.
\begin{figure*}
    \centering
    \caption{Qualitative Case Studies Illustrating the Effects of Layer Zeroing on LLaVA-Next's Reasoning. The figure* presents three comparative examples of the model's reasoning process before (base model) and after the intervention of a specific task\mbox{-}interfering layer.}
    \includegraphics[width=0.8\linewidth]{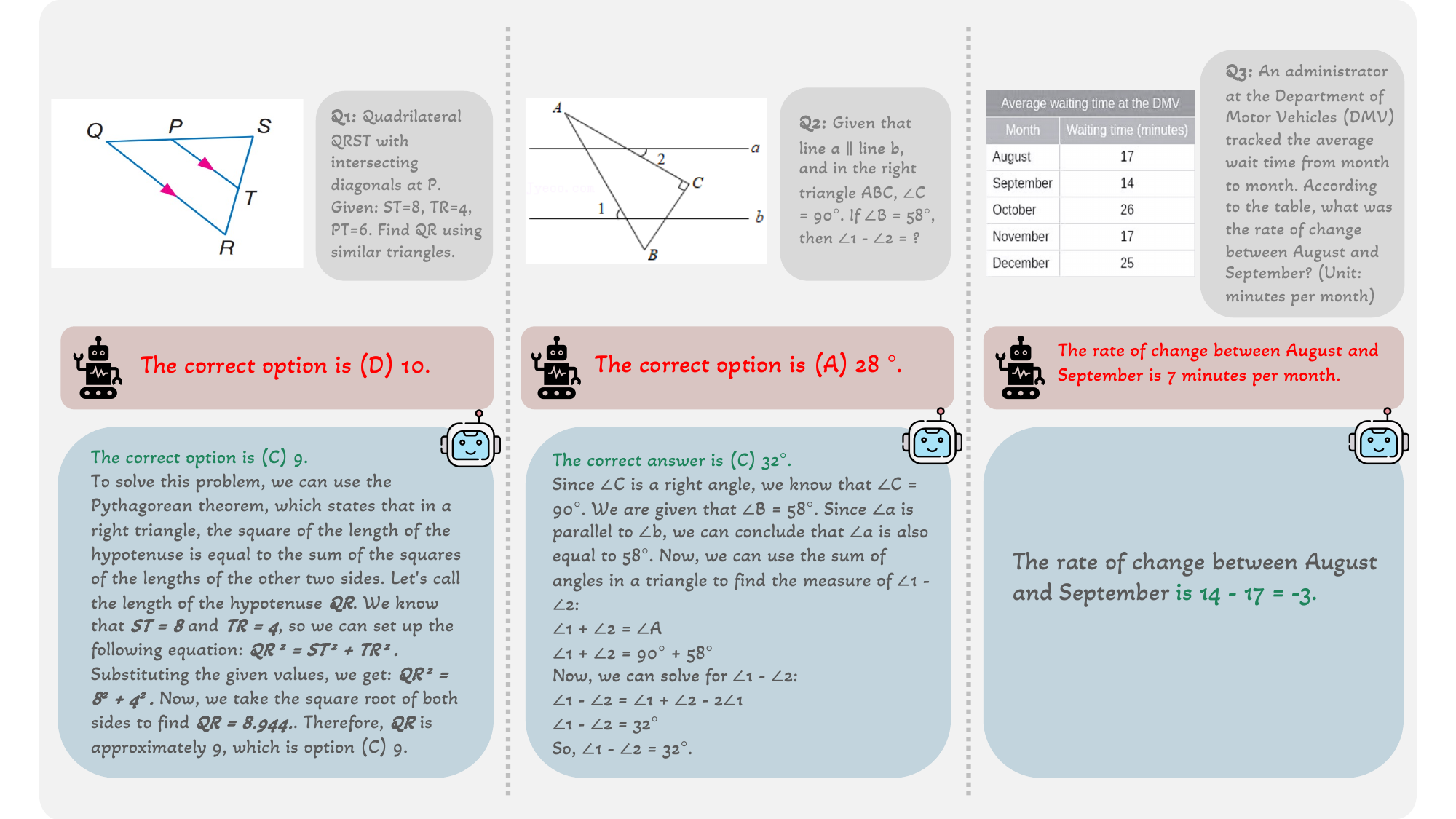}
    \label{fig6}
\end{figure*}

\begin{table*}
\centering
\renewcommand{\arraystretch}{1.3} % 调整行距
\caption{Additional results of TaLo on \textit{LLaVA-Next-LLaMA3-8B}. Here,  \textit{future prediction} is annotated as F-P, and \textit{function reasoning} is annotated as F-R.}
\scalebox{0.75}{
\begin{tabular}{l|l|ccc|cccc}
\toprule[1.2pt]
\multicolumn{1}{c|}{\multirow{2}{*}{\textbf{Model}}} & \multirow{2}{*}{\textbf{Shots}} 
& \multicolumn{3}{c|}{\textbf{MMBench}} 
& \multicolumn{4}{c}{\textbf{ScienceQA}} \\ 
\cmidrule(lr){3-5} \cmidrule(lr){6-9}
& & \textbf{F-P} & \textbf{OCR} & \textbf{F-R} & \textbf{Ecological interactions} & \textbf{The Americas: Geography} & \textbf{Oceania: Geography} & \textbf{Geography}\\ 
\midrule
\multicolumn{1}{l|}{\multirow{3}{*}{\textbf{LLaVA}}} 
& \textbf{10 shots} &43.3 \textcolor{red}{\scriptsize$\downarrow$3.3} &69.8 \textcolor{red}{\scriptsize$\downarrow$2.6}&68.6 \textcolor{gray}{\scriptsize$-$0.0} &17.6 \textcolor{gray}{\scriptsize$-$0.0} &30.0 \textcolor{mygreen}{\scriptsize$\uparrow$20.0} &21.9 \textcolor{gray}{\scriptsize$-$0.0} &41.7 \textcolor{mygreen}{\scriptsize$\uparrow$8.4}  \\
\multicolumn{1}{l|}{} & \textbf{15 shots} &48.3 \textcolor{red}{\scriptsize$\downarrow$1.1} &71.2 \textcolor{gray}{\scriptsize$-$0.0} &68.0 \textcolor{red}{\scriptsize$\downarrow$1.3} &29.4 \textcolor{mygreen}{\scriptsize$\uparrow$11.8} &25.0 \textcolor{mygreen}{\scriptsize$\uparrow$5.0} &25.0 \textcolor{mygreen}{\scriptsize$\uparrow$3.1} &39.6 \textcolor{mygreen}{\scriptsize$\uparrow$4.2} \\
\multicolumn{1}{l|}{} & \textbf{20 shots} &58.6 \textcolor{mygreen}{\scriptsize$\uparrow$6.9} &78.8 \textcolor{mygreen}{\scriptsize$\uparrow$3.8} &71.0\textcolor{mygreen}{\scriptsize$\uparrow$2.2} &29.4\textcolor{mygreen}{\scriptsize$\uparrow$11.8} &35.0 \textcolor{mygreen}{\scriptsize$\uparrow$20.0} &15.6\textcolor{gray}{\scriptsize$-$0.0} &33.3 \textcolor{mygreen}{\scriptsize$\uparrow$2.1}  \\ 
\bottomrule[1.1pt]
\end{tabular}
}
\label{tab:llava}
\end{table*}

\begin{table*}
\centering
\caption{Additional results of TaLo on \textit{Qwen2-VL-2B}, where \textit{structuralized imagetext understanding} is annotated as S-I-U, \textit{attribute recognition} is annotated as AR, \textit{physical relation} is annotated as PR, and \textit{celebrity recognition} is annotated as CR.}
\renewcommand{\arraystretch}{1.3} % 调整行距
\scalebox{0.7}{
\begin{tabular}{l|l|cccc|ccccc}
\toprule[1.2pt]
\multicolumn{1}{c|}{\multirow{2}{*}{\textbf{Model}}} & \multirow{2}{*}{\textbf{Shots}} 
& \multicolumn{4}{c|}{\textbf{MMBench}} 
& \multicolumn{5}{c}{\textbf{ScienceQA}} \\ 
\cmidrule(lr){3-6} \cmidrule(lr){7-11}
& & \textbf{S-I-U} & \textbf{AR} & \textbf{PR} & \textbf{CR} 
  & \textbf{Astronomy} & \textbf{The Americas: Geography} & \textbf{Genes to traits} & \textbf{Solutions} & \textbf{Force and motion} \\ 
\midrule
\multicolumn{1}{l|}{\multirow{3}{*}{\textbf{Qwen-VL}}} 
& \textbf{10 shots} & 52.5 \textcolor{mygreen}{\scriptsize$\uparrow$3.3} & 72.3 \textcolor{mygreen}{\scriptsize$\uparrow$0.9} & 47.6 \textcolor{mygreen}{\scriptsize$\uparrow$1.6} & 75.0 \textcolor{mygreen}{\scriptsize$\uparrow$2.0}
& 35.5 \textcolor{mygreen}{\scriptsize$\uparrow$3.2} & 25.0 \textcolor{mygreen}{\scriptsize$\uparrow$15.0} & 21.9 \textcolor{mygreen}{\scriptsize$\uparrow$12.5} & 24.4 \textcolor{mygreen}{\scriptsize$\uparrow$2.2} & 35.3 \textcolor{mygreen}{\scriptsize$\uparrow$5.9} \\

\multicolumn{1}{l|}{} & \textbf{15 shots} & 52.3 \textcolor{mygreen}{\scriptsize$\uparrow$3.2} & 72.1 \textcolor{mygreen}{\scriptsize$\uparrow$2.0} & 50.8 \textcolor{mygreen}{\scriptsize$\uparrow$3.2} & 75.3 \textcolor{mygreen}{\scriptsize$\uparrow$1.5} 
& 38.7 \textcolor{mygreen}{\scriptsize$\uparrow$6.4} & 10.0 \textcolor{gray}{\scriptsize$-$0.0} & 25.0 \textcolor{mygreen}{\scriptsize$\uparrow$3.1} & 28.9 \textcolor{mygreen}{\scriptsize$\uparrow$8.9} & 29.4 \textcolor{red}{\scriptsize$\downarrow$5.9} \\

\multicolumn{1}{l|}{} & \textbf{20shots} & 48.5 \textcolor{mygreen}{\scriptsize$\uparrow$2.5} & 73.4 \textcolor{mygreen}{\scriptsize$\uparrow$1.1} & 55.6 \textcolor{mygreen}{\scriptsize$\uparrow$3.2} & 76.6 \textcolor{mygreen}{\scriptsize$\uparrow$1.0} 
& 32.3 \textcolor{mygreen}{\scriptsize$\uparrow$6.5} & 15.0 \textcolor{mygreen}{\scriptsize$\uparrow$5.0} & 28.1 \textcolor{mygreen}{\scriptsize$\uparrow$6.2} & 35.6 \textcolor{mygreen}{\scriptsize$\uparrow$15.6} & 52.9 \textcolor{mygreen}{\scriptsize$\uparrow$17.6} \\ 
\bottomrule[1.1pt]
\end{tabular}
}

\label{tab:qwen}
\end{table*}

\begin{table*}[h]
\centering
\renewcommand{\arraystretch}{1.5} % 增加行间距
\small
\scalebox{1}{
\begin{tabular}{c|p{.7\textwidth}}
\toprule
\multicolumn{2}{c}{\textbf{Cluster Category and Tasks Included}} \\ \midrule
\textbf{Cluster 1 (Quantitative Reasoning)} & \ttfamily numeric commonsense, arithmetic reasoning, geometry reasoning, algebraic reasoning, geometry problem solving, math word problem, figure* question answering, statistical reasoning, Cities, Informational texts: level 1, Particle motion and energy \\
\addlinespace

\textbf{Cluster 2 (Analytical Reasoning)} & \ttfamily image\_emotion, biology, engineering, public health, instance reasoning, math, geography, visual reasoning, architecture \& engineering, diagnostics \& laboratory medicine, electronics, psychology, maps, magnets, plant reproduction, domain-specific vocabulary, genes to traits \\
\addlinespace

\textbf{Cluster 3 (Scientific Knowledge)} & \ttfamily scientific reasoning, textbook question answering, chemistry, medicine, economics, physics, sociology, art \& design, science \& technology, astronomy, plants, weather, fossils, thermal energy, natural resources, pharmacy, humanities \& social science, literature \\
\addlinespace

\textbf{Cluster 4 (Integrative Reasoning)} & \ttfamily logical reasoning, accounting, history, pharmacy, engineering practices, ecology, world religions, persuasive strategies, adaptations and natural selection, economics, sociology, humanities, design, literature, natural science theory, political history \\
\addlinespace

\textbf{Cluster 5 (Perceptual Categorization)} & \ttfamily action recognition, attribute recognition, image quality, coarse perception, fine-grained perception, object localization, classification, ecosystems, force and motion, solutions, states of matter, scene understanding, OCR, image\_scene, visual elements, structuralized image-text understanding, attribute comparison, celebrity recognition \\
\addlinespace

\textbf{Cluster 6 (Predictive Reasoning)} & \ttfamily future prediction, identity reasoning, spatial relationship, electronics, psychology, math, music, plant reproduction, velocity \& acceleration, instance interaction, physical geography, classification and scientific names, context clues, text understanding \\
\addlinespace

\textbf{Cluster 7 (Relational Understanding)} & \ttfamily nature\_relation, physical\_relation, social relation, art theory, civil war and reconstruction, age of exploration, ancient Mesopotamia, plate tectonics, geology, animals, agriculture, ecosystems, cultural history, world religions, state capitals \\
\bottomrule
\end{tabular}
}
\caption{Clusters and their included tasks from various benchmarks.}
\label{tab:clusters_tasks}
\end{table*}

\subsection{Empirical Validations of Task\mbox{-}Interfering \\ Layers}
\label{results}

\begin{figure*}[h]
    \centering
    \begin{subfigure}[b]{0.8\textwidth}
        \includegraphics[width=\textwidth]{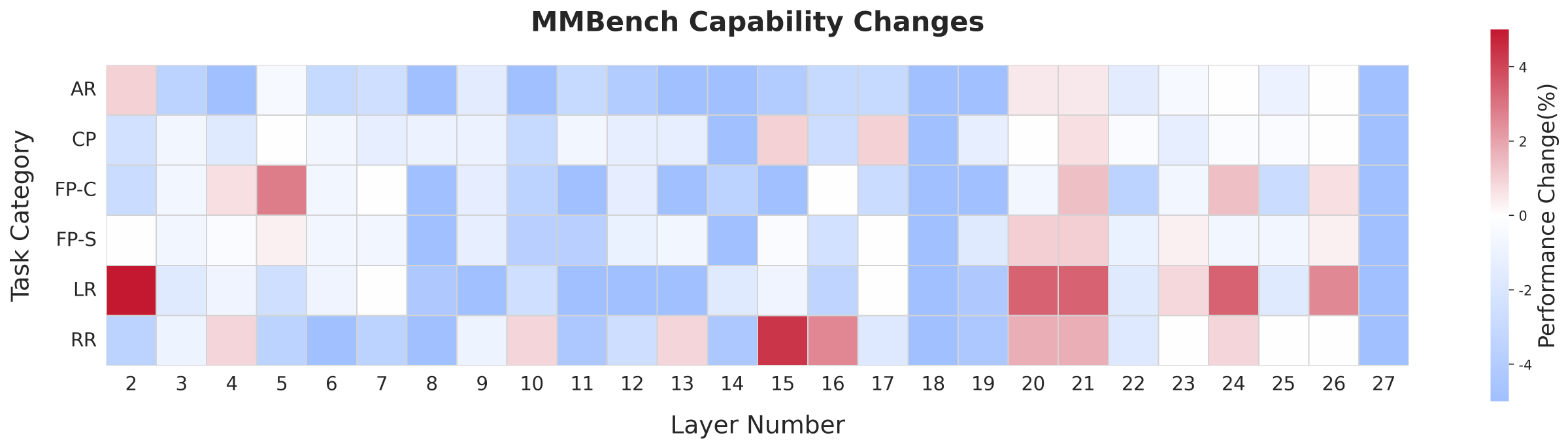}
        \caption{Accuracy change heatmap on Qwen2-VL.}
        \label{mmbench:image1}
    \end{subfigure}
    \vspace{1em} % Vertical space between subfigures
    \begin{subfigure}[b]{0.8\textwidth}
        \includegraphics[width=\textwidth]{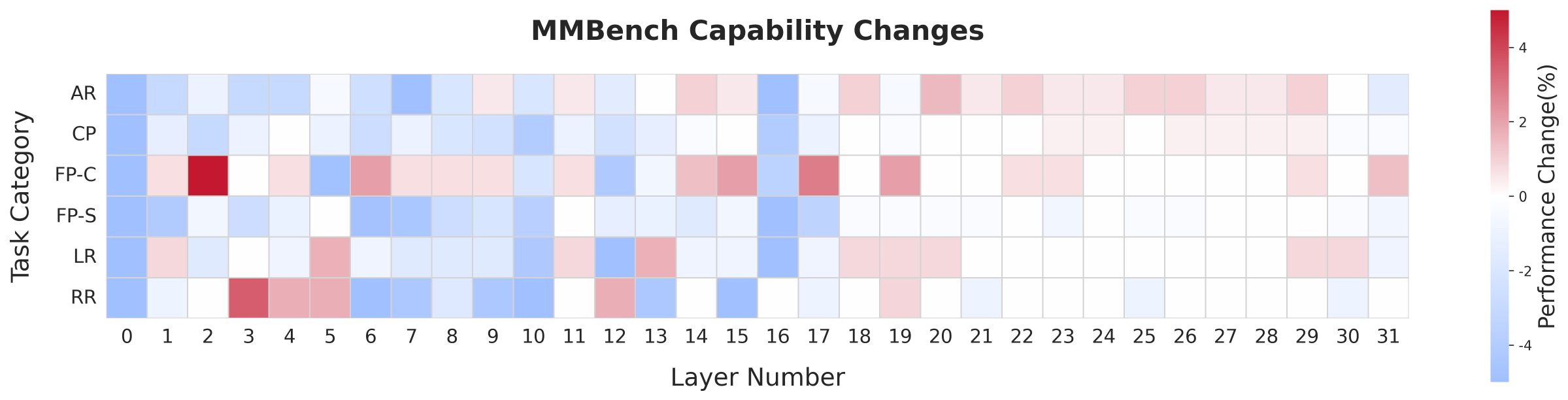}
        \caption{Accuracy change heatmap on LLaVA-Next.}
        \label{mmbench:image2}
    \end{subfigure}
    \vspace{1em}
    \begin{subfigure}[b]{0.8\textwidth}
        \includegraphics[width=\textwidth]{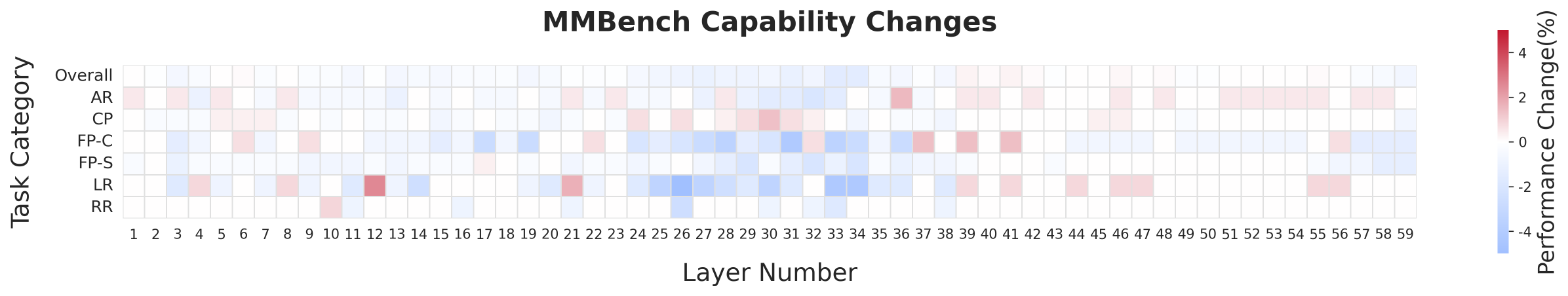}
        \caption{Accuracy change heatmap on InternVL2.}
        \label{fig:image3}
    \end{subfigure}
    \caption{Accuracy change heatmaps on MMBench (Uniform Scaling).}
    \label{fig:mmbench_heatmaps}
\end{figure*}

\begin{figure*}[h]
    \centering
    \includegraphics[width=0.7\linewidth]{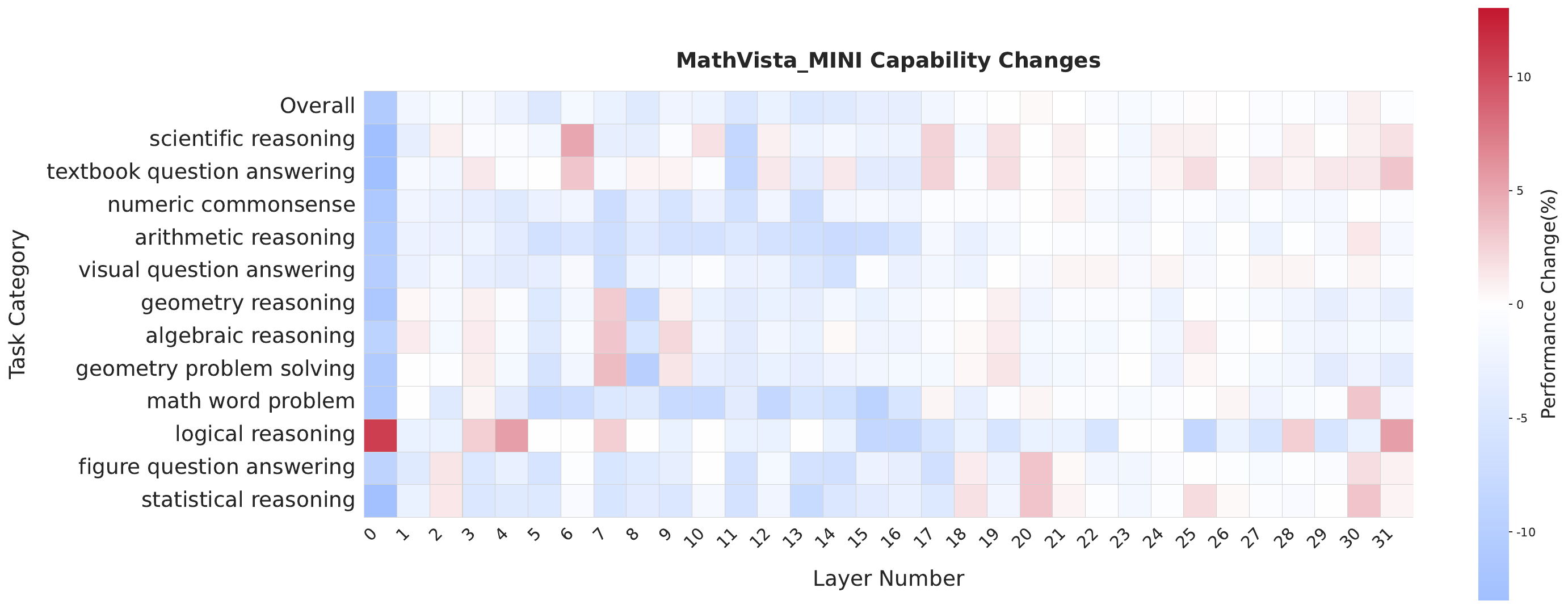}
    \caption{Accuracy change heatmap for LLaVA-Next on MathVista-MINI (Uniform Scaling).}
    \label{fig:mathvista}
\end{figure*}

\begin{figure*}[h]
    \centering
    \includegraphics[width=0.7\linewidth]{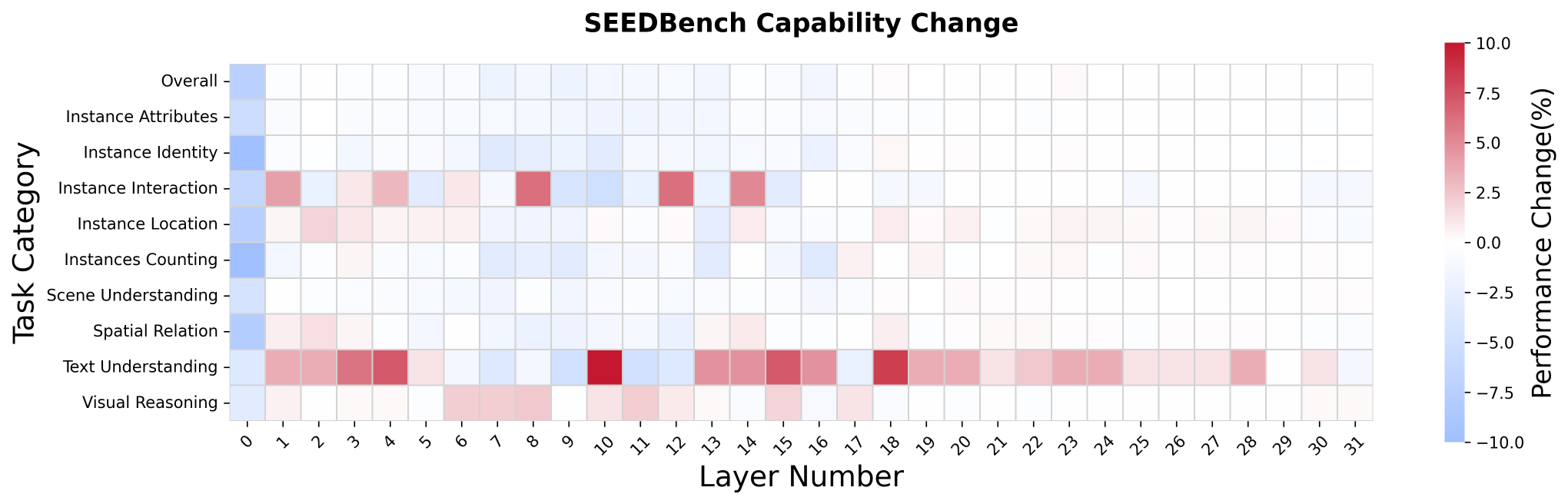}
    \caption{Accuracy change heatmap for LLaVA-Next on SEEDBench (Uniform Scaling).}
    \label{fig:seedbench}
\end{figure*}

\begin{figure*}[htbp]
    \centering
    \includegraphics[width=0.7\linewidth]{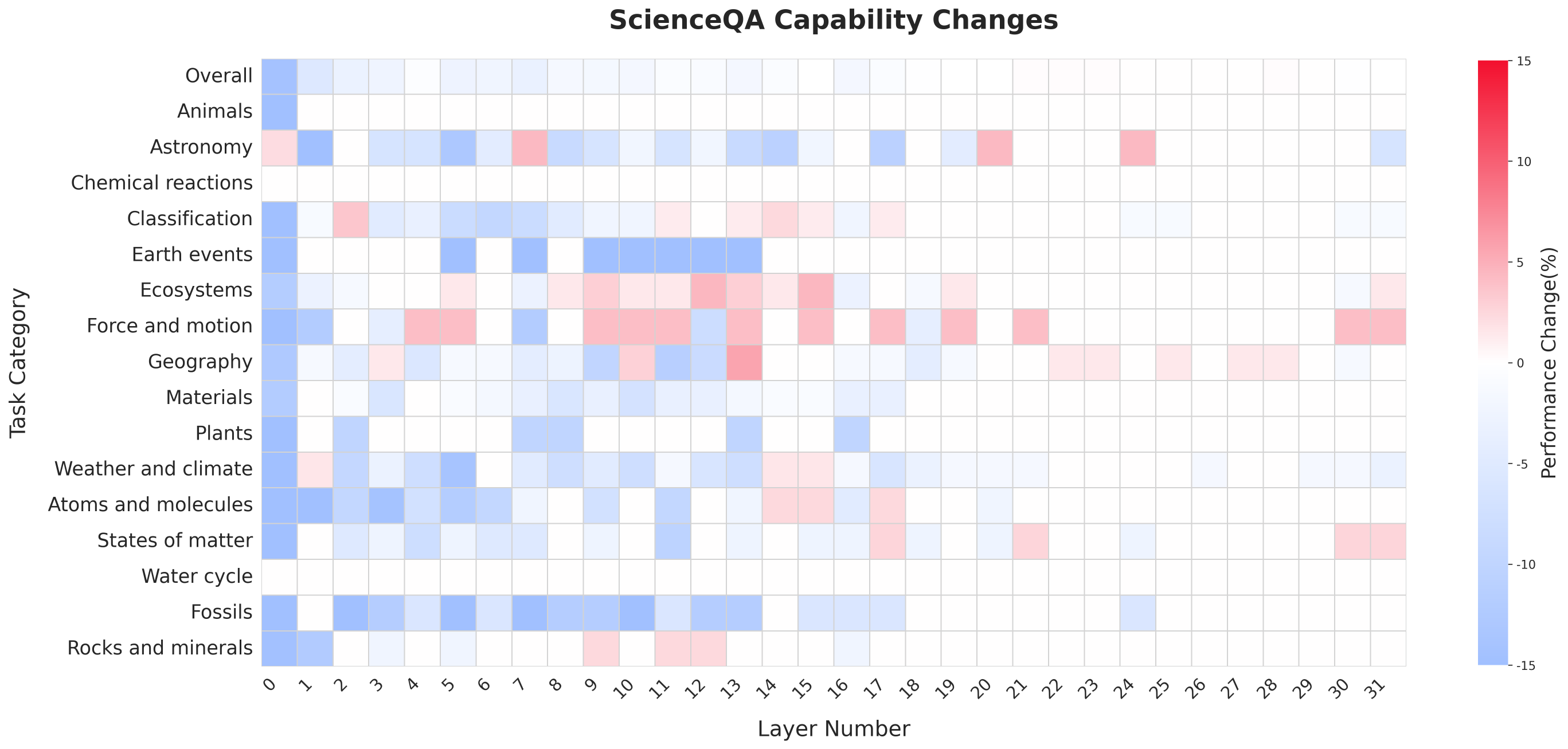}
    \caption{Accuracy change heatmap for LLaVA-Next on ScienceQA (Uniform Scaling).}
    \label{fig:scienqa}
\end{figure*}

\begin{figure*}[htbp]
    \centering
    \begin{subfigure}[b]{0.9\textwidth}
        \includegraphics[width=\textwidth]{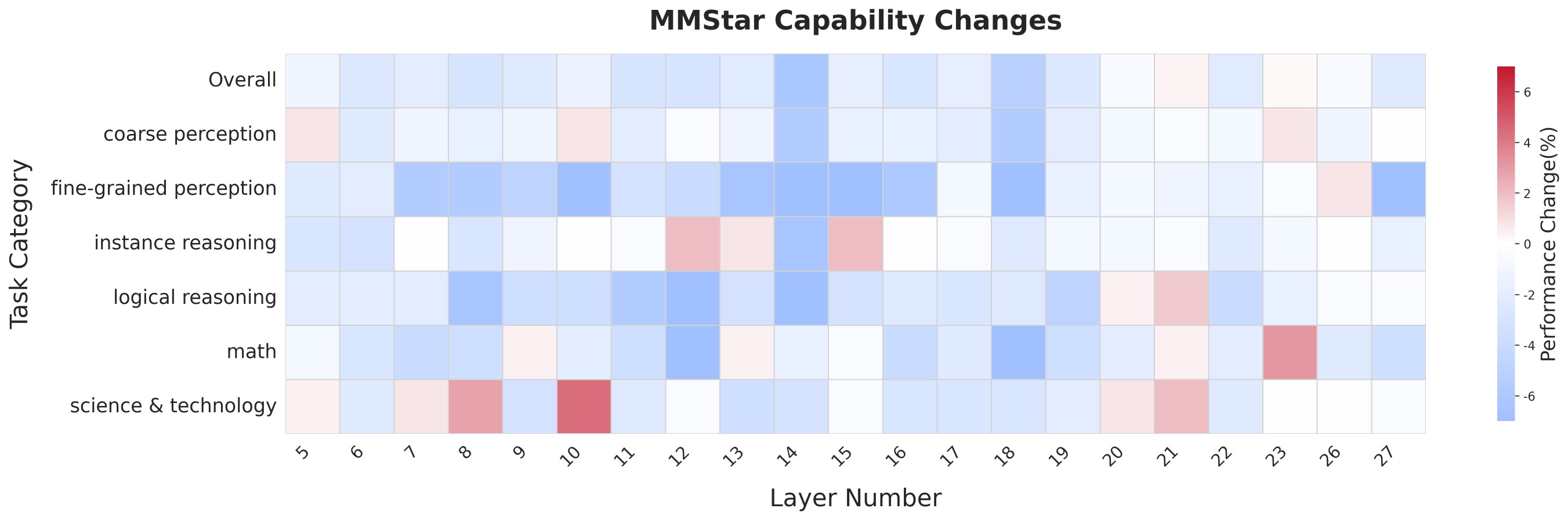}
        \caption{Accuracy change heatmap on Qwen2-VL.}
        \label{mmstar:image1}
    \end{subfigure}
    
    \vspace{5em}
    
    \begin{subfigure}[b]{0.9\textwidth}
        \includegraphics[width=\textwidth]{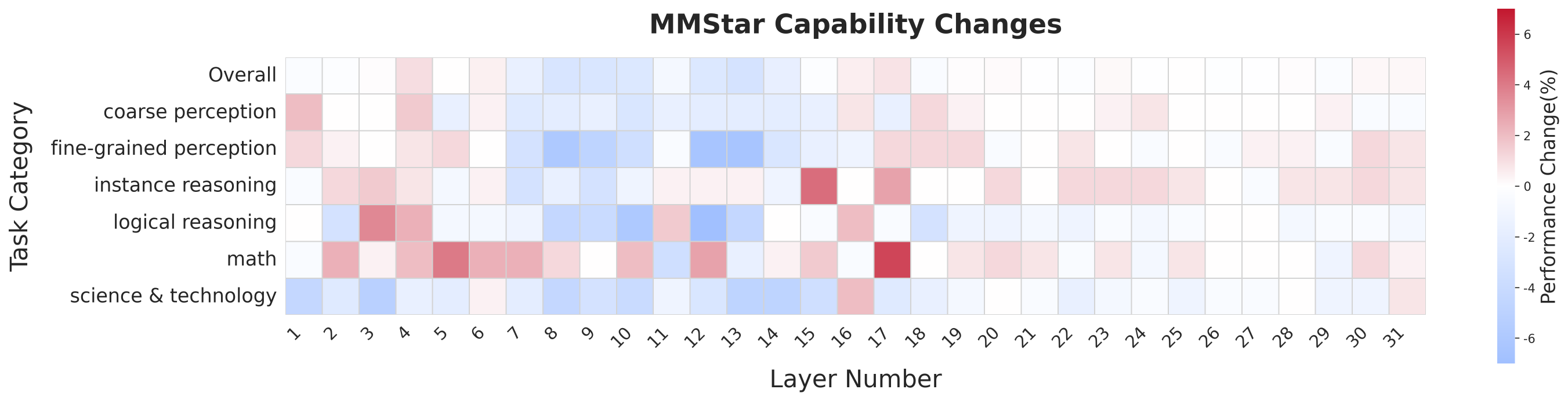}
        \caption{Accuracy change heatmap on LLaVA-Next.}
        \label{mmstar:image2}
    \end{subfigure}
    
    \vspace{5em}
    
    \begin{subfigure}[b]{0.9\textwidth}
        \includegraphics[width=\textwidth]{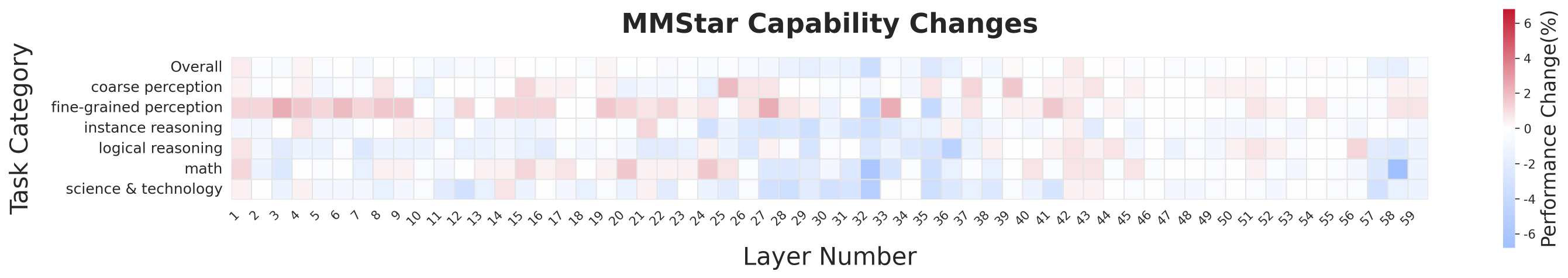}
        \caption{Accuracy change heatmap on InternVL2.}
        \label{mmstar:image3}
    \end{subfigure}
    \caption{Accuracy change heatmaps on MMStar (Uniform Scaling).}
    \label{fig:mmstar_heatmaps}
\end{figure*}

\begin{figure*}[h]
    \centering
    \begin{subfigure}[b]{1\textwidth}
    \centering
        \includegraphics[width=0.72\textwidth]{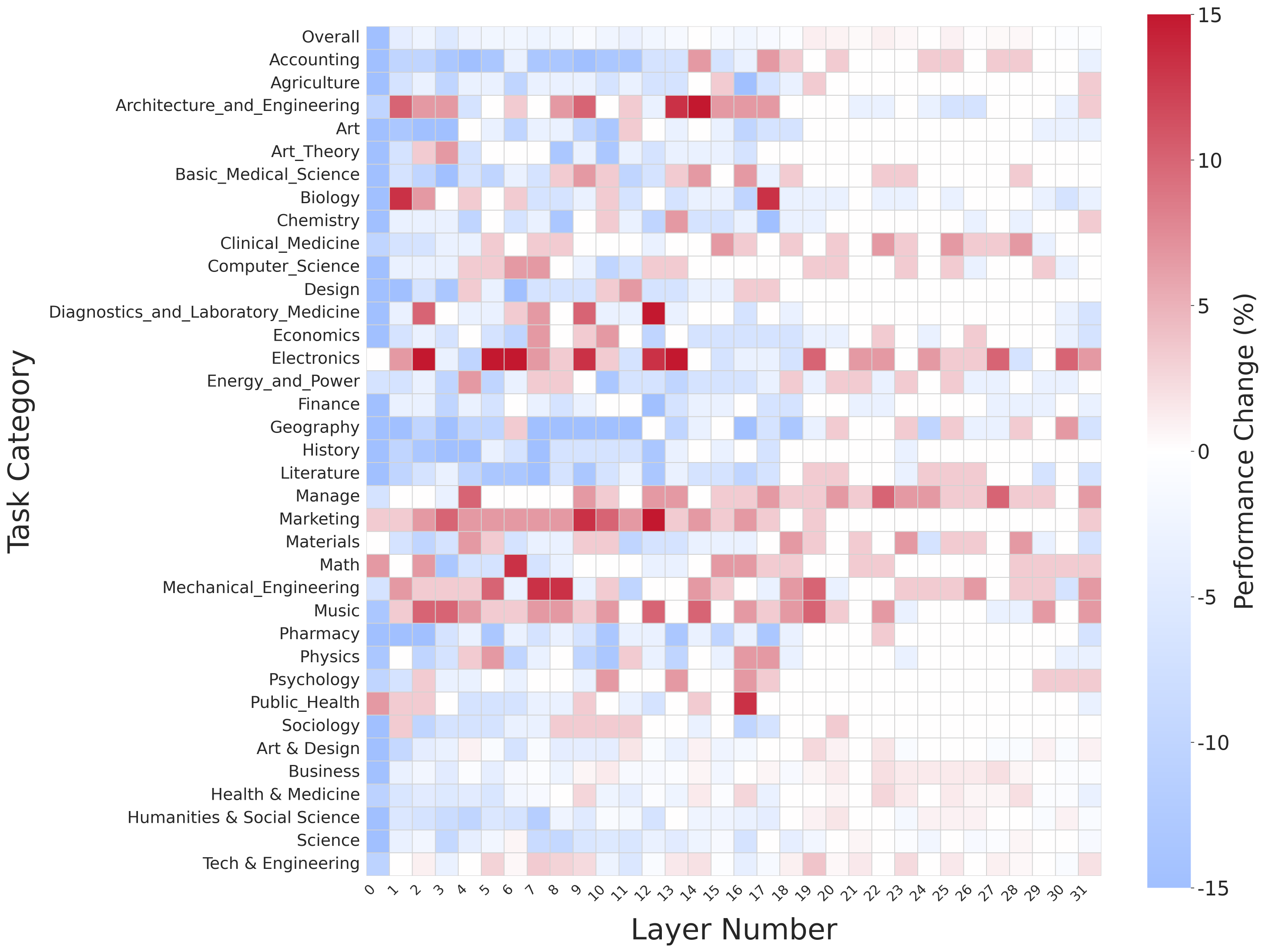}
        \caption{Accuracy change heatmap on LLaVA-Next.}
        \label{mmmu:image2}
    \end{subfigure}
    
    \vspace{1em}
    
    \begin{subfigure}[b]{0.7\textwidth}
        \includegraphics[width=\textwidth]{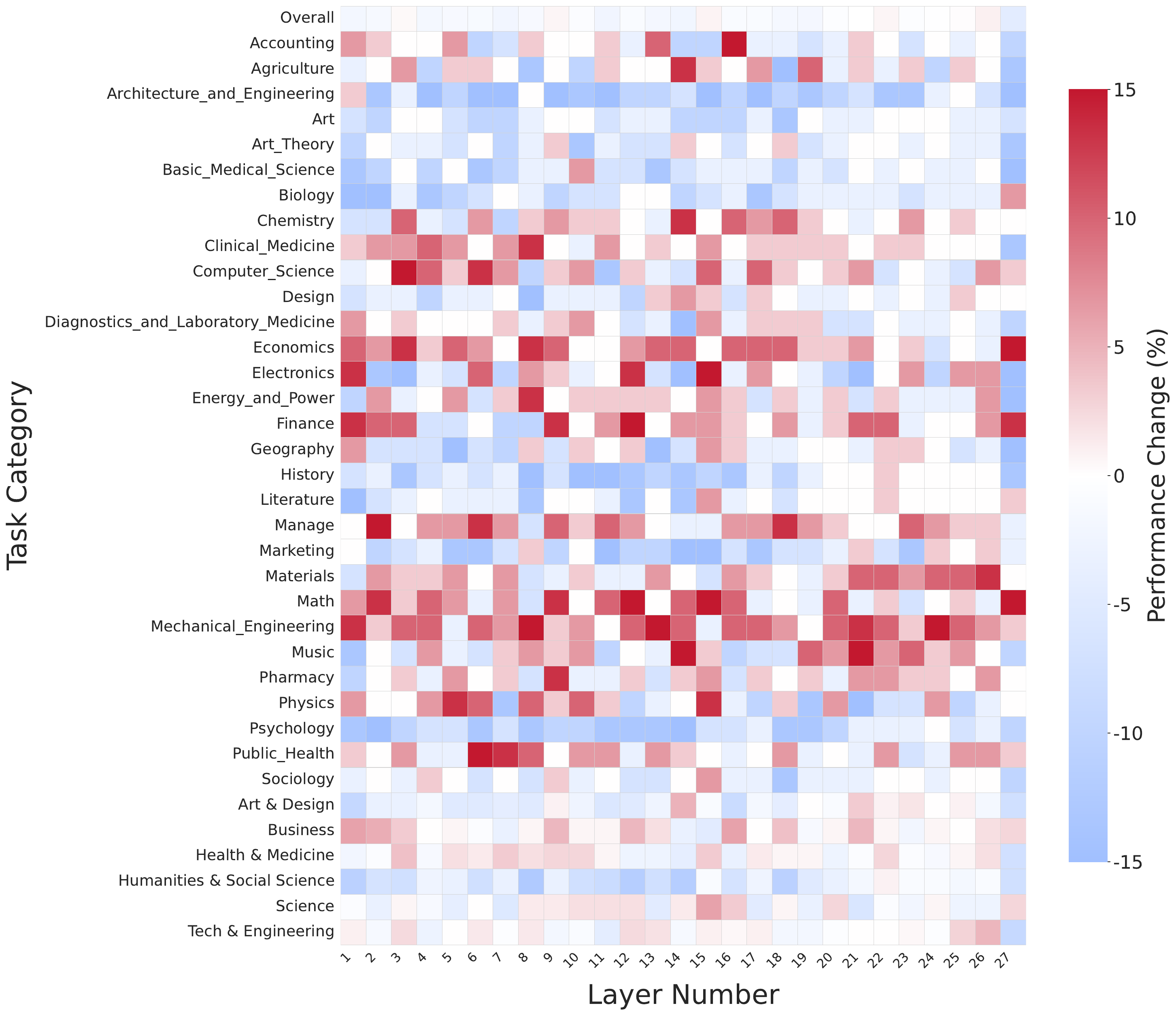}
        \caption{Accuracy change heatmap on Qwen2-VL.}
        \label{mmmu:image3}
    \end{subfigure}
    \caption{Accuracy change heatmaps on MMMU (Uniform Scaling).}
    \label{fig:mmmu_heatmaps}
\end{figure*}

\begin{figure*}[h]
    \centering
    \begin{subfigure}[b]{0.8\textwidth}
        \includegraphics[width=\textwidth]{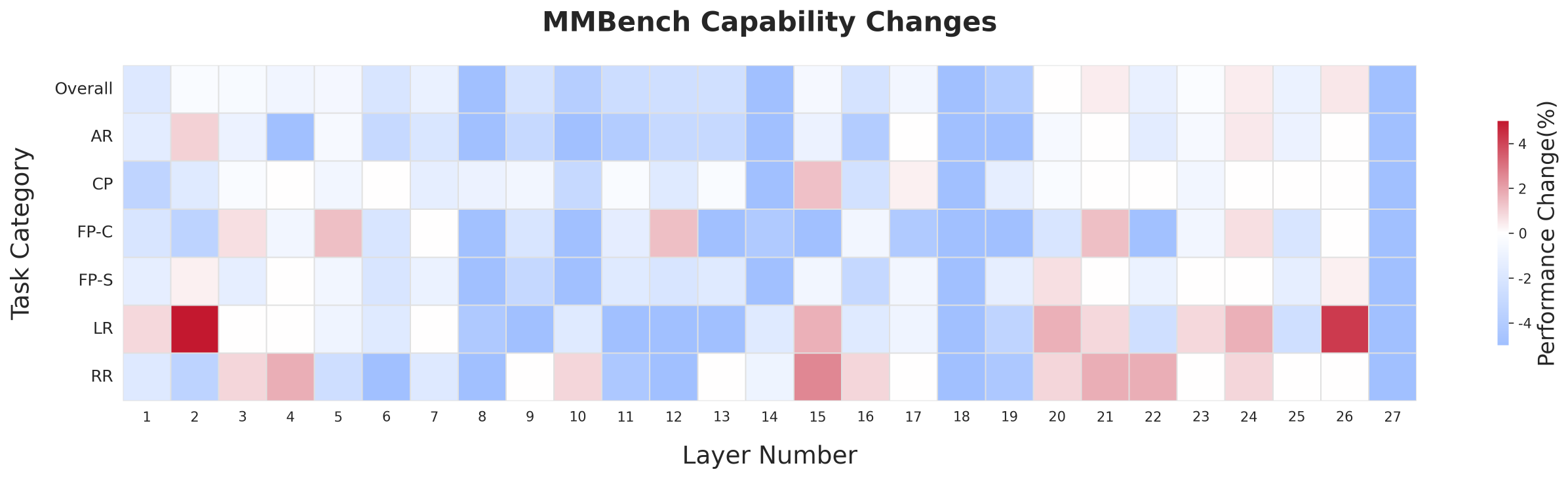}
        \caption{Accuracy change heatmap on Qwen2-VL.}
        \label{mmbench:image10}
    \end{subfigure}
    \vspace{1em} % Vertical space between subfigures
    \begin{subfigure}[b]{0.8\textwidth}
        \includegraphics[width=\textwidth]{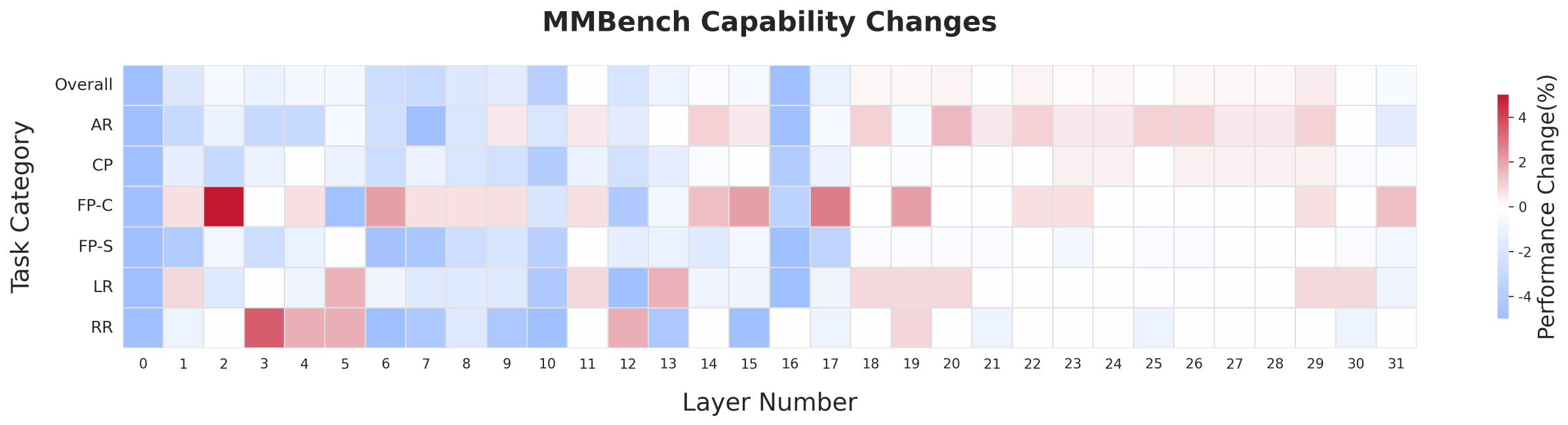}
        \caption{Accuracy change heatmap on LLaVA-Next.}
        \label{mmbench:image20}
    \end{subfigure}
    \vspace{1em}
    \begin{subfigure}[b]{0.8\textwidth}
        \includegraphics[width=\textwidth]{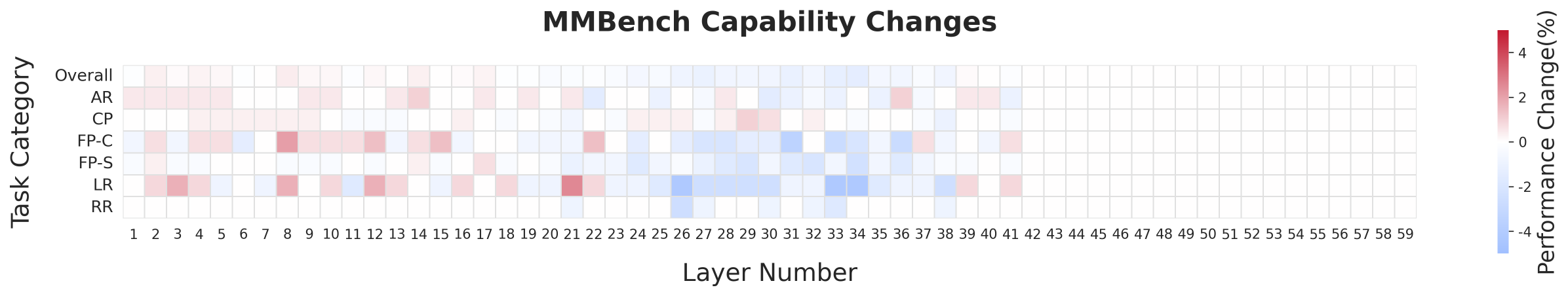}
        \caption{Accuracy change heatmap on InternVL2.}
        \label{fig:image30}
    \end{subfigure}
    \caption{Accuracy change heatmaps on MMBench (Zeroing).}
    \label{fig:mmbench0}
\end{figure*}

\begin{figure*}[h]
    \centering
    \includegraphics[width=0.8\linewidth]{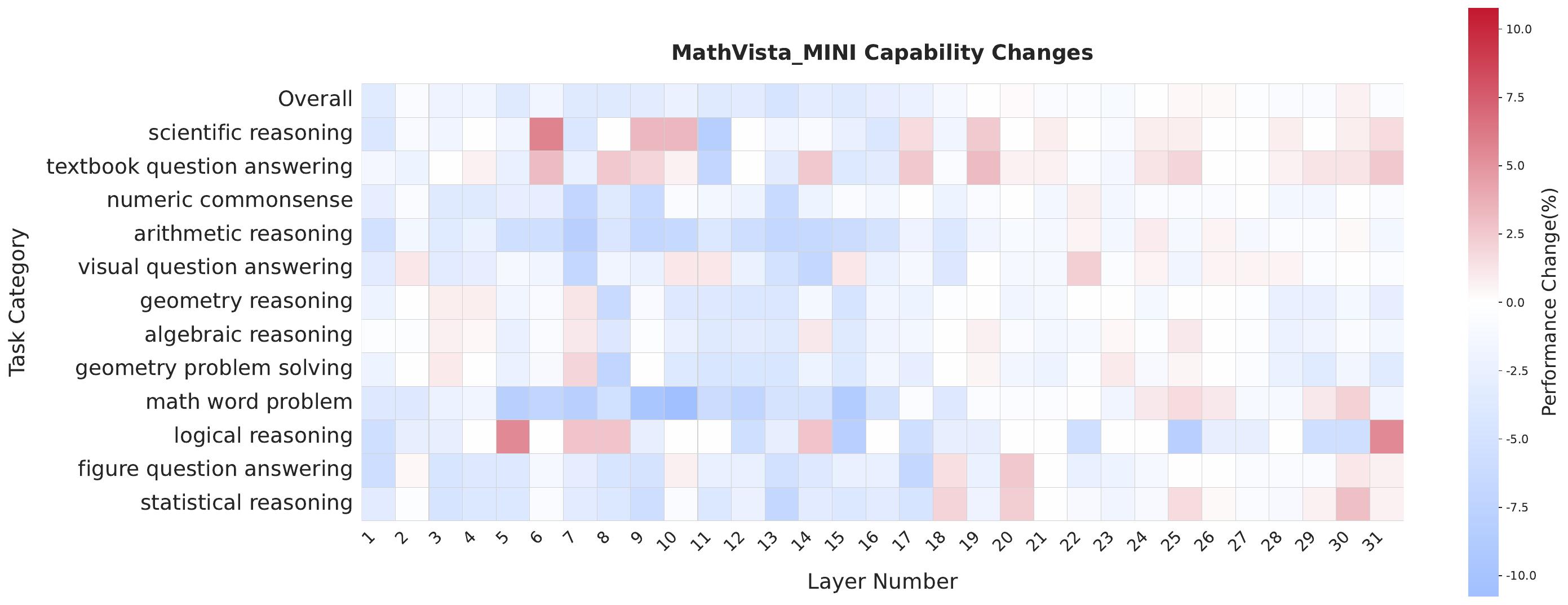}
    \caption{Accuracy change heatmap for LLaVA-Next on MathVista-MINI (Zeroing).}
    \label{fig:mathvista0}
\end{figure*}

\begin{figure*}[h]
    \centering
    \includegraphics[width=0.8\linewidth]{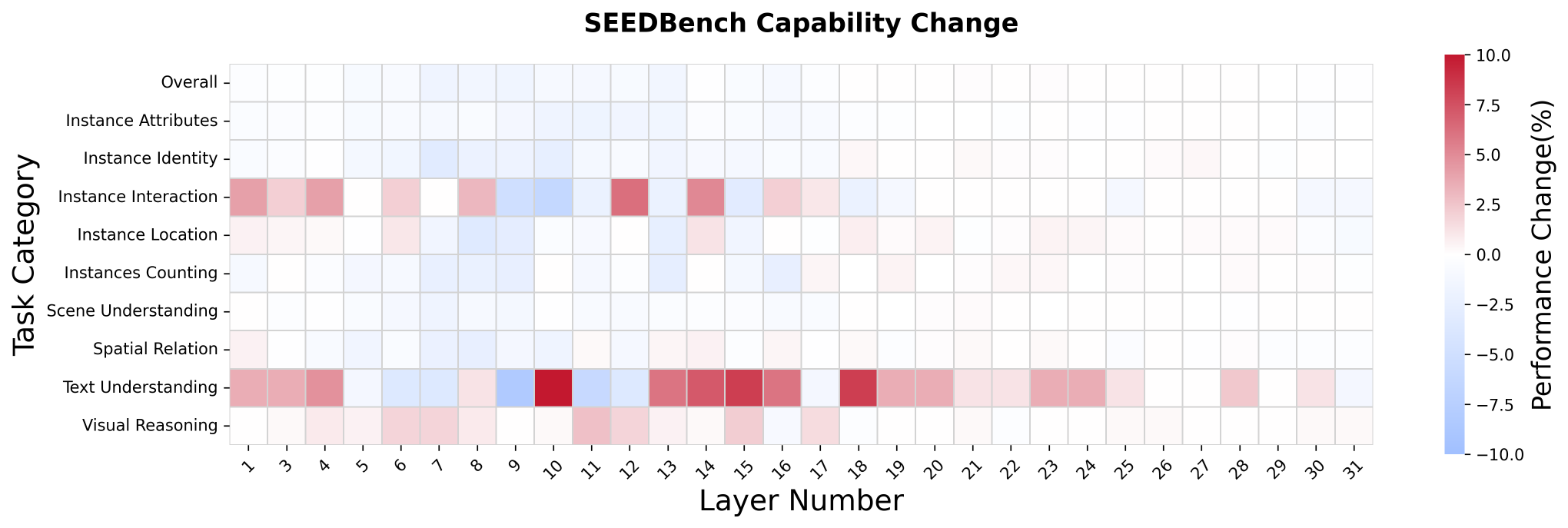}
    \caption{Accuracy change heatmap for LLaVA-Next on SEEDBench (Zeroing).}
    \label{fig:seed0}
\end{figure*}

\begin{figure*}[h]
    \centering
    \includegraphics[width=0.8\linewidth]{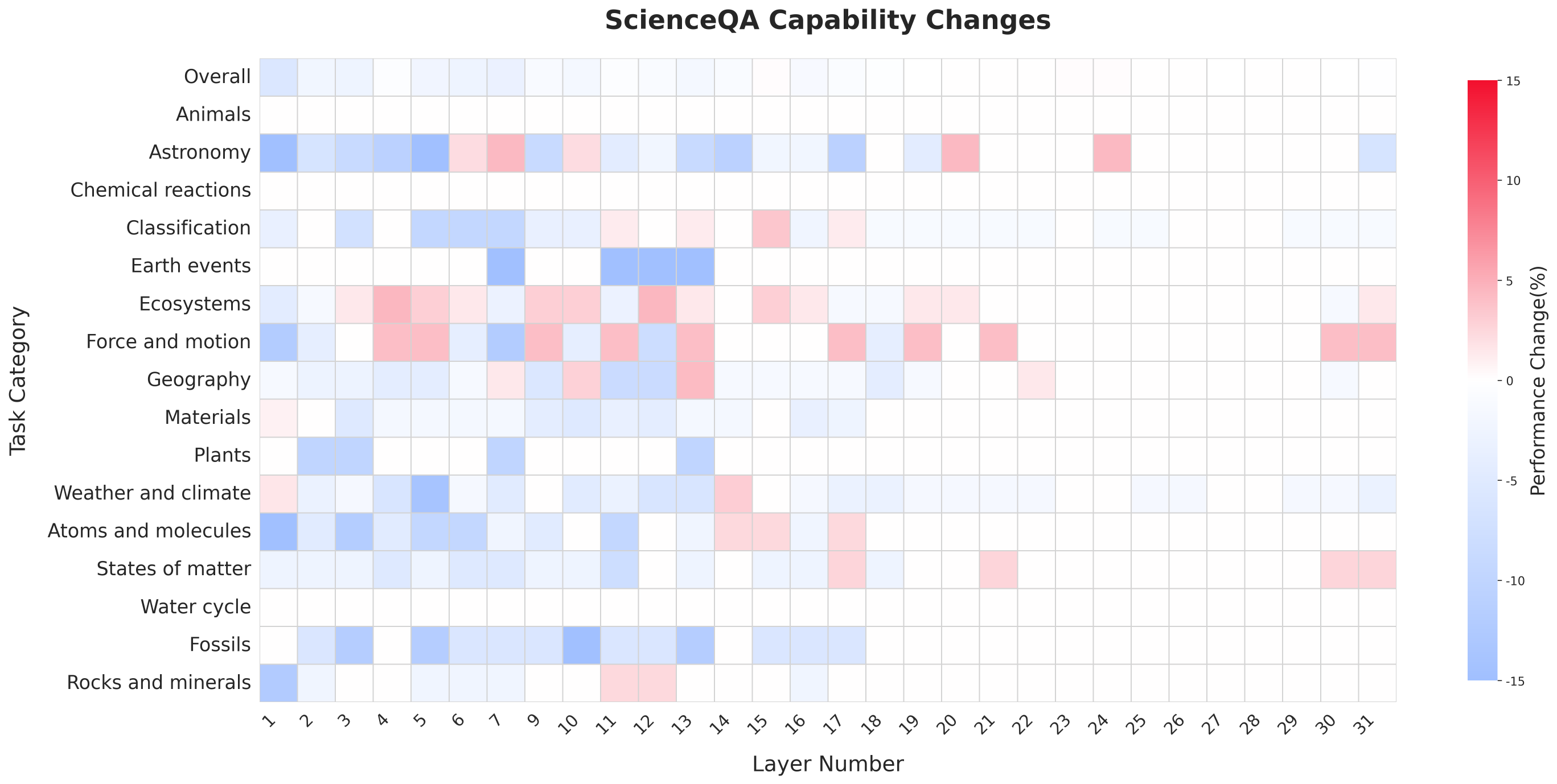}
    \caption{Accuracy change heatmap for LLaVA-Next on ScienceQA (Zeroing).}
    \label{fig:sciencqa0}
\end{figure*}

\begin{figure*}[h]
    \centering
    \begin{subfigure}[b]{0.8\textwidth}
        \includegraphics[width=\textwidth]{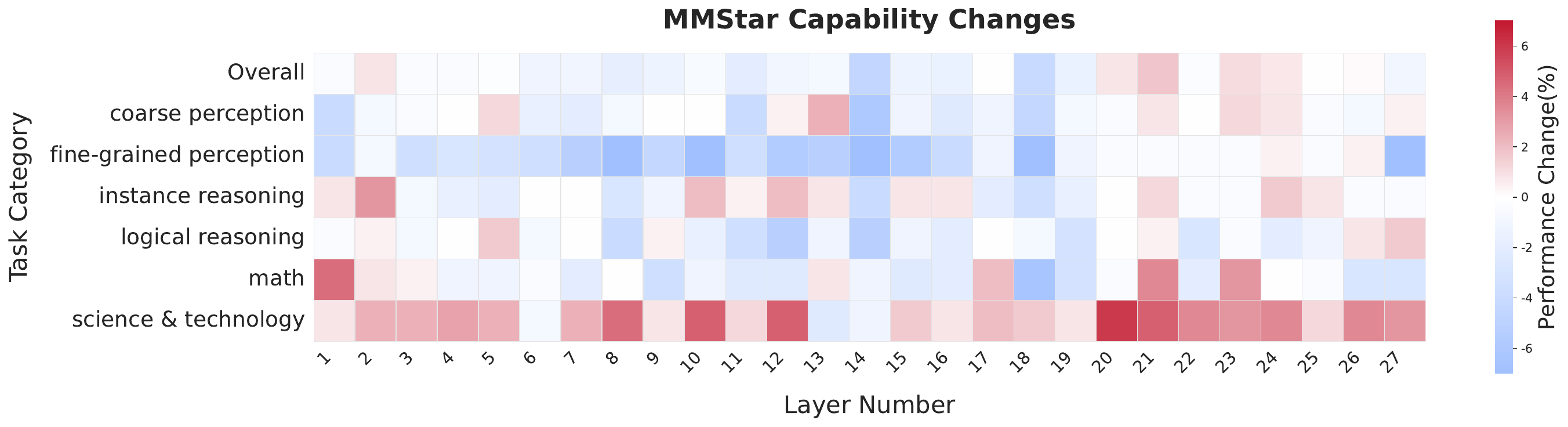}
        \caption{Accuracy change heatmap on Qwen2-VL.}
        \label{mmstar:image1}
    \end{subfigure}
    \vspace{1em}
    \begin{subfigure}[b]{0.8\textwidth}
        \includegraphics[width=\textwidth]{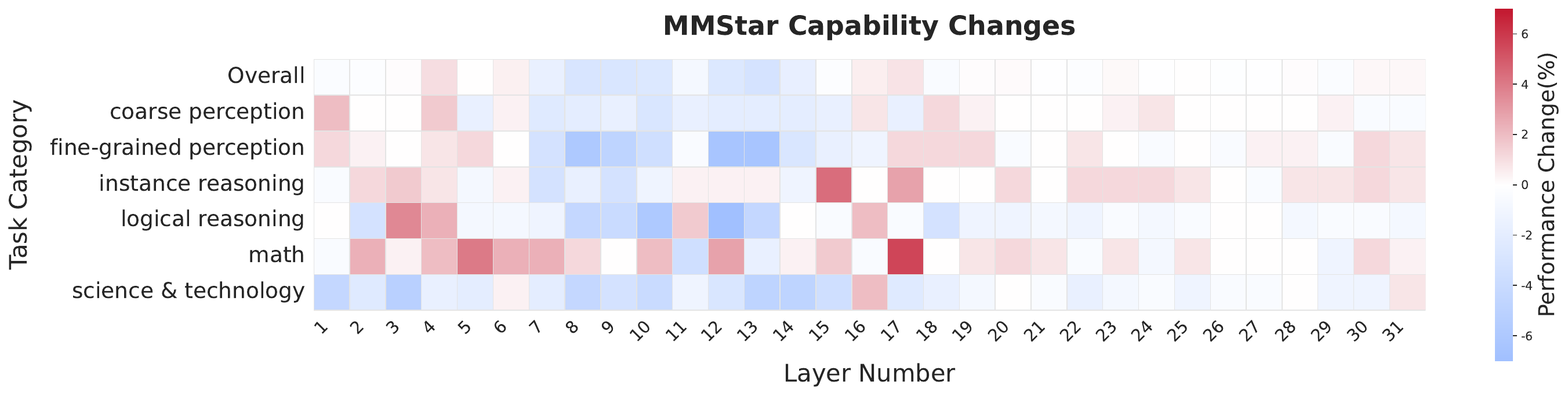}
        \caption{Accuracy change heatmap on LLaVA-Next.}
        \label{mmstar:image2}
    \end{subfigure}
    \vspace{1em}
    \begin{subfigure}[b]{0.8\textwidth}
        \includegraphics[width=\textwidth]{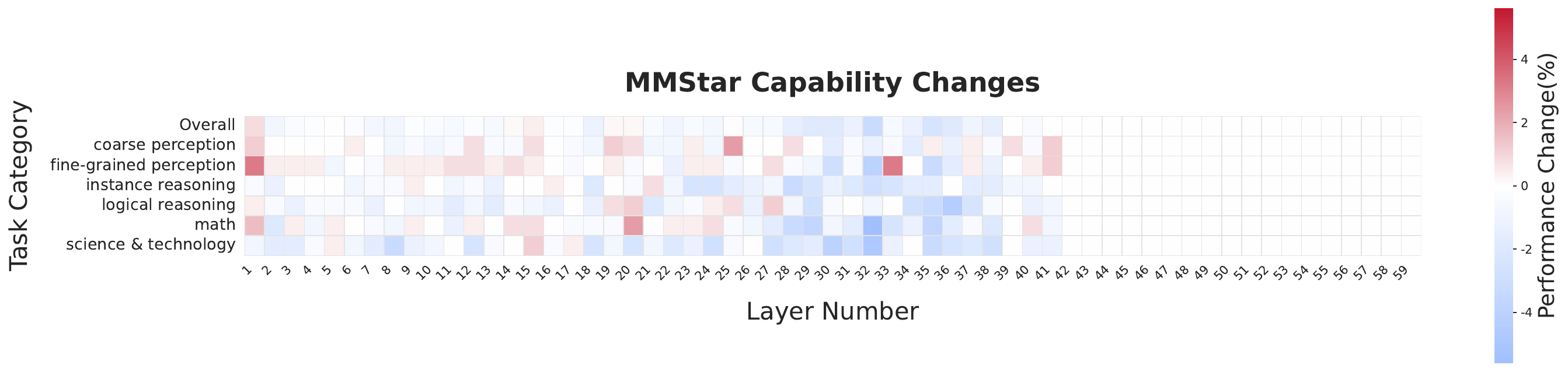}
        \caption{Accuracy change heatmap on InternVL2.}
        \label{mmstar:image3}
    \end{subfigure}
    \caption{Accuracy change heatmaps on MMStar (Zeroing).}
    \label{fig:mmstar_0}
\end{figure*}

\begin{figure*}[h]
    \centering
    \begin{subfigure}[b]{0.75\textwidth}
        \includegraphics[width=\textwidth]{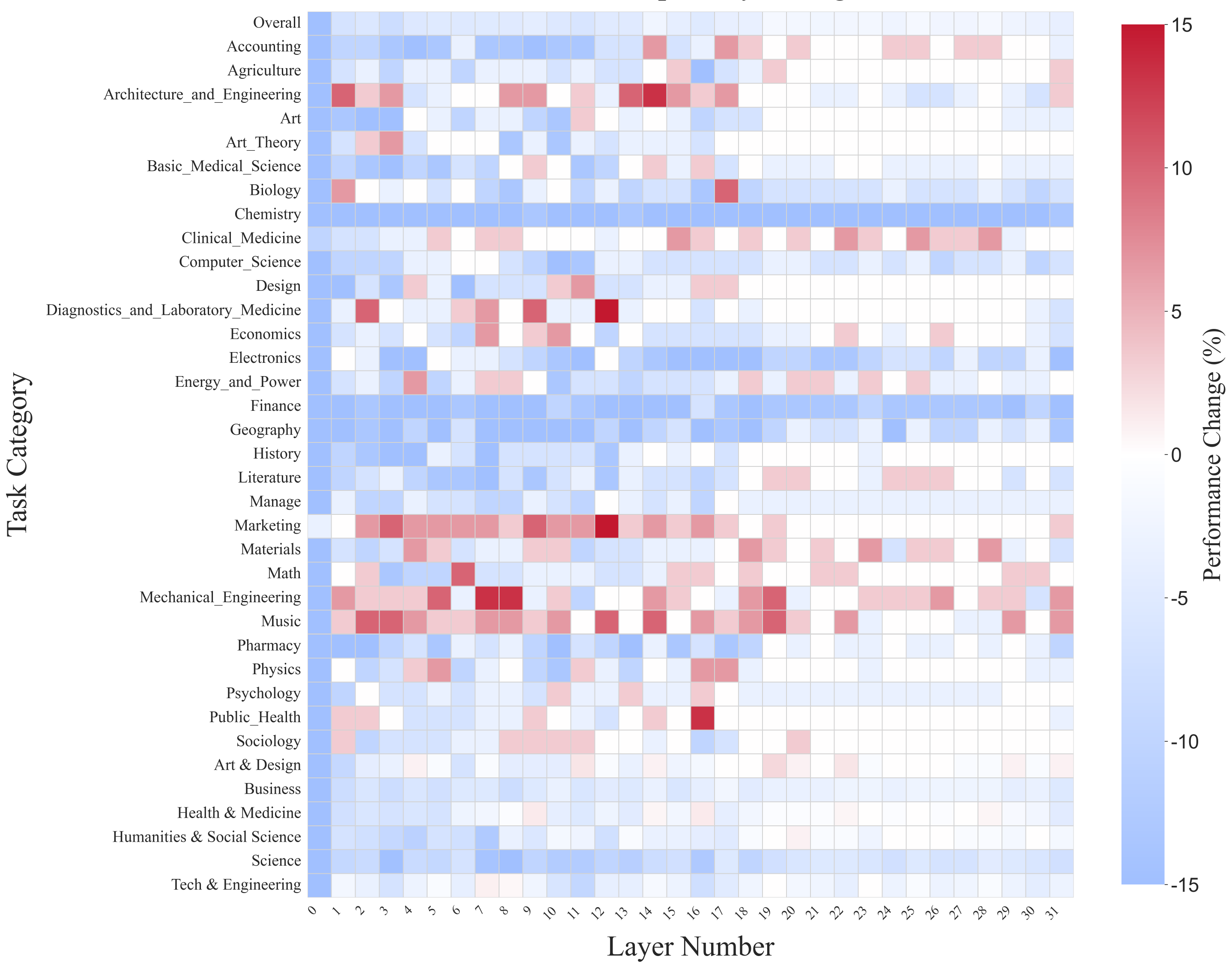}
        \caption{Accuracy change heatmap on LLaVA-Next.}
        \label{mmmu:image02}
    \end{subfigure}
    
    \vspace{1em}
    
    \begin{subfigure}[b]{0.7\textwidth}
        \includegraphics[width=\textwidth]{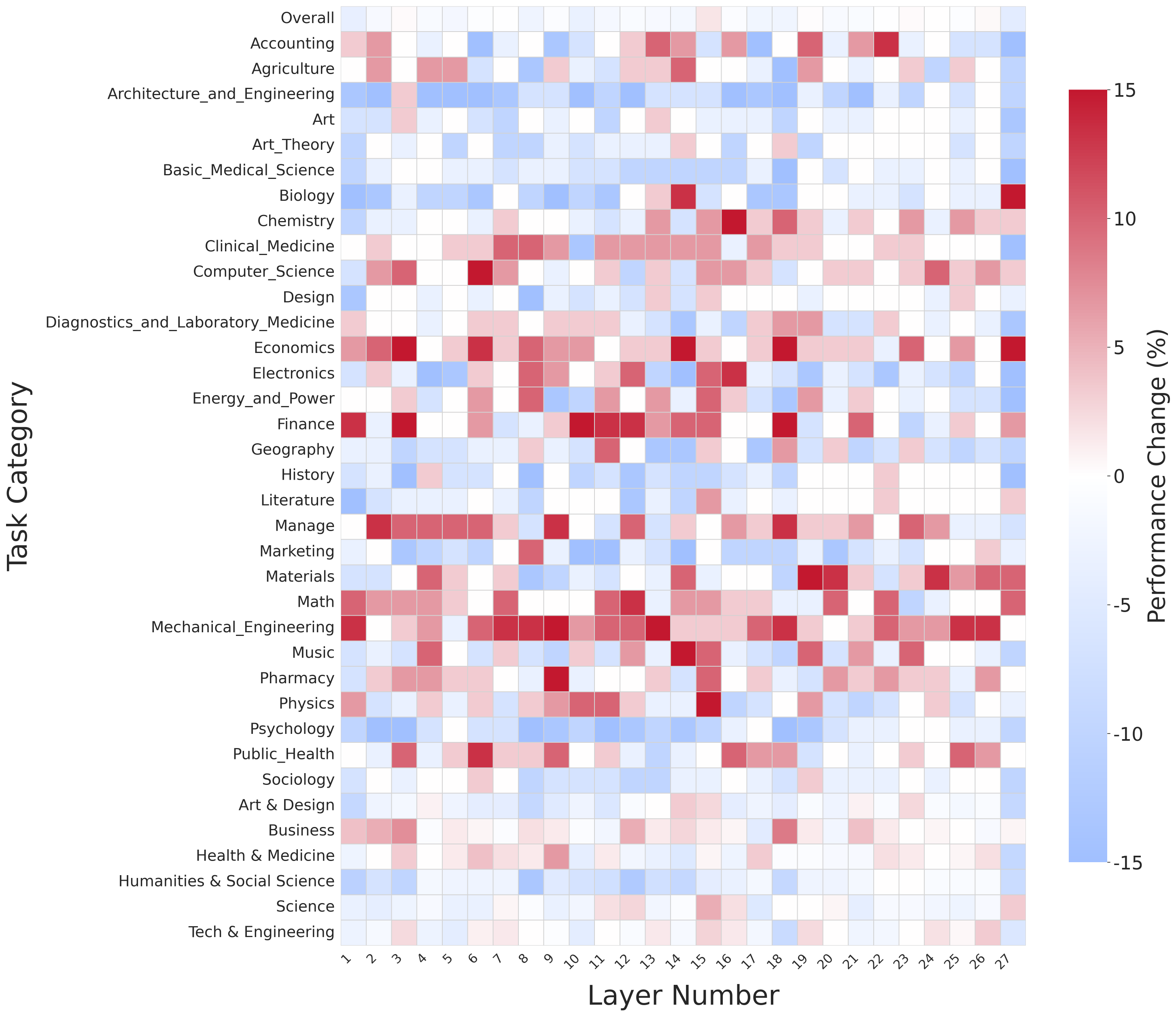}
        \caption{Accuracy change heatmap on Qwen2-VL.}
        \label{mmmu:image03}
    \end{subfigure}
    \caption{Accuracy change heatmaps on MMMU (Zeroing).}
    \label{fig:mmmu_heatmaps0}
\end{figure*}

The accuracy change heatmaps across multiple models (\textit{LLaVA-Next-Llama3-8B}, \textit{Qwen2-VL-2B}, and \textit{InternVL2-40B}) under different intervention strategies are shown in Figures~\ref{fig:mmbench_heatmaps} to \ref{fig:mmmu_heatmaps0}. Evaluated on diverse benchmarks, these heatmaps reveal consistent patterns of layer-specific performance gains, forming the core empirical basis for the task\mbox{-}interfering layers phenomenon. Rather than isolated anomalies, the results suggest a systemic trade-off in how individual layers support competing task demands, observable across model scales and architectures.

We observe that Task-Layer Interaction Vectors exhibit significant morphological differences across VLMs, even for identical tasks. We argue this is not a limitation but a critical, expected finding. Since foundation models diverge in architectures, pre-training data, and objectives, they develop unique paths of internal functional specialization. Consequently, task interference is inherently a model-specific phenomenon; a layer hindering performance in one model may not do so in another. This variability renders universal, static removal strategies impractical. It strongly underscores the necessity of our adaptive approach, TaLo. By utilizing a small probing set, TaLo effectively handles this heterogeneity, dynamically identifying and bypassing the specific interfering layers unique to each model's representational space.

\end{document}